\documentclass{article}
\pdfoutput=1  

\usepackage{microtype}
\usepackage{graphicx}
\usepackage{subfigure}
\usepackage{booktabs} 

\usepackage{hyperref}  
\usepackage{url}

\usepackage[accepted]{icml2023}

\usepackage{amsmath}
\usepackage{amssymb}
\usepackage{mathtools}
\usepackage{amsthm}

\usepackage[capitalize,noabbrev]{cleveref}

\theoremstyle{plain}

\theoremstyle{definition}

\theoremstyle{remark}

\usepackage[textsize=tiny]{todonotes}

\usepackage{listings}
\usepackage{wrapfig}
\usepackage{floatflt}
\usepackage{multirow}
\usepackage{float}

\newcommand{\norm}[1]{\left\lVert#1\right\rVert}

\usepackage{xcolor}  
\usepackage{colortbl}  

\definecolor{gray}{rgb}{0.5,0.5,0.5}
\definecolor{darkerblue}{rgb}{0,0.08,0.45}
\definecolor{darkergreen}{RGB}{21, 152, 56}
\definecolor{darkerred}{RGB}{220, 35, 120}
\definecolor{gray94}{gray}{.94}
\definecolor{gray90}{gray}{.90}
\definecolor{gray85}{gray}{.85}

\newcommand{\ul}{\underline}

\newcommand{\gray}[1]{\textcolor{gray}{#1}}

\usepackage{pifont}%
\usepackage{xspace}
\usepackage{placeins}
\newcommand{\xmarkg}{\textcolor{gray}{\ding{55}}\xspace}%

\icmltitlerunning{Architecture-Agnostic Masked Image Modeling -- From ViT back to CNN}

\begin{document}

\twocolumn[
\icmltitle{Architecture-Agnostic Masked Image Modeling -- From ViT back to CNN}



\icmlsetsymbol{equal}{*}


\begin{icmlauthorlist}
\icmlauthor{Siyuan Li}{equal,west,zju}
\icmlauthor{Di Wu}{equal,west,zju}
\icmlauthor{Fang Wu}{west,tsu}
\icmlauthor{Zelin Zang}{west,zju}
\icmlauthor{Stan Z. Li}{west}
\end{icmlauthorlist}


\icmlaffiliation{west}{AI Lab, Research Center for Industries of the Future, Westlake University, Hangzhou, 310000, China}
\icmlaffiliation{zju}{College of Computer Science and Technology, Zhejiang University, Hangzhou, 310000, China}
\icmlaffiliation{tsu}{Institute of AI Industry Research, Tsinghua University, Beijing, 100084, China}

\icmlcorrespondingauthor{Stan Z. Li}{stan.z.li@westlake.edu.cn}

\icmlkeywords{Self-supervised learning, Masked Image Modeling, Pre-training}

\vskip 0.3in
]



\printAffiliationsAndNotice{\icmlEqualContribution} 

\begin{abstract}
Masked image modeling, an emerging self-supervised pre-training method, has shown impressive success across numerous downstream vision tasks with Vision transformers. Its underlying idea is simple: a portion of the input image is masked out and then reconstructed via a pre-text task. However, the working principle behind MIM is not well explained, and previous studies insist that MIM primarily works for the Transformer family but is incompatible with CNNs.
In this work, we observe that MIM essentially teaches the model to learn better middle-order interactions among patches for more generalized feature extraction. We then propose an Architecture-Agnostic Masked Image Modeling framework (A$^2$MIM), which is compatible with both Transformers and CNNs in a unified way. Extensive experiments on popular benchmarks show that A$^2$MIM learns better representations without explicit design and endows the backbone model with the stronger capability to transfer to various downstream tasks.
\end{abstract}

\section{Introduction}
\label{sec:intro}
Supervised deep learning with large-scale annotated data has witnessed an explosion of success in computer vision (CV)~\citep{NIPS2012alexnet, he2016deep} and natural language processing (NLP)~\citep{vaswani2017attention}. However, a large number of high-quality annotations are not always available in real-world applications. Learning representations without supervision by leveraging pre-text tasks has become increasingly popular.

In CV, early self-supervised learning approaches \citep{eccv2016coloring, iccv2015relativeloc,iclr2018rotation} aim to capture invariant features through predicting transformations applied to the same image. However, these methods rely on vision ad-hoc heuristics, and the learned representations are less generic. Recently, contrastive learning approaches~\citep{eccv2020CMC,chen2020simple,cvpr2020moco} have witnessed significant progress, even outperforming supervised methods on several downstream tasks~\citep{2020mocov2,nips2020byol,zbontar2021barlow}. More recently, inspired by masked autoencoding methods~\citep{devlin2018bert, Radford2018GPT1} in NLP, Masked Image Modeling (MIM) methods~\citep{bao2021beit,he2021masked,wei2021masked, xie2021simmim} have brought about new advances for self-supervised pre-training on CV tasks. The transition from human language understanding to NLP masked autoencoding is quite natural because the filling of missing words in a sentence requires comprehensive semantic understanding. In analogy, humans can understand and imagine masked content by visually filling the missing structures in an image containing occluded parts.

Different from contrastive learning, which yields a clustering effect by pulling similar samples and pushing away dissimilar samples, MIM pre-training methods have not been extensively explored in the context of the expected knowledge learned. Existing works mainly focus on improving downstream tasks performance via explicit design such as trying different prediction targets~\citep{wei2021masked}, adopting pre-trained tokenizer~\citep{zhou2021ibot}, utilizing complex Transformer decoder~\citep{he2021masked} or combining with contrastive learning~\citep{el2021large}. Moreover, the success of existing MIM methods is largely confined to Vision Transformer (ViT) structures~\citep{dosovitskiy2020image} since it leads to under-performing performance to directly apply mask token~\citep{devlin2018bert} and positional embedding to CNNs.


In this work, we carry out systematic experiments and show that MIM as a pre-training task essentially teaches the model to learn better middle-order interactions between patches for more generalized feature extraction regardless of the underlying network structure. Compared to the local texture features learned by low-order interactions between patches, more complex features such as shape and edge could be extracted via middle-order interactions among patches. The interaction of patches could be considered as information fusion via both the convolution operation of a CNN and the self-attention mechanism of a Transformer. That is to say, CNN and Transformer should both benefit from  better middle-order interactions with MIM as the pre-text task.

To bridge the gap of MIM in terms of network architectures based on our extensive experimental analysis, we propose an Architecture-Agnostic Masked Image Modeling framework (A$^2$MIM) that focuses on enhancing the middle-order interaction capabilities of the network. Specifically, we mask the input image with the mean RGB value and place the mask token at intermediate feature maps of the network. In addition, we propose a loss in the Fourier domain to further enhance the middle-order interaction capability of the network. Our contributions are summarized as follows:
\begin{itemize}
    \vspace{-0.5em}
    \item We conducted systematic experiments and showed the essence of MIM is to better learn middle-order interactions between patches but not reconstruction quality.
    \vspace{-1.5em}
    \item We proposed a novel MIM-based framework dubbed A$^2$MIM that bridges the gap between CNNs and Transformers. We are also the first to perform MIM on CNNs without adopting designs native to ViTs that outperform contrastive learning counterparts.
    \vspace{-0.25em}
    \item Extensive experiments with both Transformers and CNNs on ImageNet-1K and public benchmarks for various downstream tasks show that our method improves performances on pre-trained representations.
\end{itemize}


\section{Related Work}
\label{sec:related}

\paragraph{Contrastive Learning.}
Contrastive learning (CL) learns instance-level discriminative representations by extracting invariant features over distorted views of the same data. MoCo~\citep{cvpr2020moco} and SimCLR~\citep{chen2020simple} adopted different mechanisms to introduce numerous negative samples for contrast with the positive. BYOL~\citep{nips2020byol} and its variants~\citep{chen2020simsiam,nips2021revitalizing} further eliminate the requirement of negative samples to avoid representation collapse. Besides pairwise contrasting, SwAV~\citep{caron2020unsupervised} clusters the data while enforcing consistency between multi-augmented views of the same image. Barlow Twins~\citep{zbontar2021barlow} and its variants~\citep{icml2021WMSE,iclr2022VICReg} proposed to measure the cross-correlation matrix of distorted views of the same image to avoid representation collapsing. Meanwhile, some efforts have been made on top of contrastive methods to improve pre-training quality for specific downstream tasks~\citep{iccv2021detco, xiao2021region,cvpr2021casting}. MoCo.V3~\citep{chen2021empirical} and DINO~\citep{iccv2021dino} adopted ViT~\citep{dosovitskiy2020image} in CL pre-training to replace CNN backbones.

\vspace{-0.25em}
\paragraph{Autoregressive Modeling.}
Autoencoders (AE) is a typical type of architecture that allows representation learning with no annotation requirement~\citep{hinton1993autoencoders}. By forcing denoising property onto the learned representations, denoising autoencoders \citep{vincent2008extracting, vincent2010stacked} are a family of AEs that reconstruct the uncorrected input signal with a corrupted version of the signal as input.
Generalizing the notion of denoising autoregressive modeling, masked predictions attracted the attention of both the NLP and CV communities. BERT~\citep{devlin2018bert} performs masked language modeling (MLM) where the task is to classify the randomly masked input tokens. Representations learned by BERT as pre-training generalize well to various downstream tasks. For CV, inpainting tasks~\citep{pathak2016context} to predict large missing regions using CNN encoders and colorization tasks~\citep{eccv2016coloring} to reconstruct the original color of images with removed color channels are proposed to learn representation without supervision. With the introduction of Vision Transformers (ViTs) \citep{dosovitskiy2020image,iccv2021Swin}, iGPT~\citep{chen2020generative} predicts succeeding pixels given a sequence of pixels as input. MAE~\citep{he2021masked} and BEiT~\citep{bao2021beit} randomly mask out input image patches and reconstruct the missing patches with ViTs. Compared to MAE, MaskFeat~\citep{wei2021masked} and SimMIM~\citep{xie2021simmim} adopt linear layers as the decoder instead of another Transformer as in MAE. MaskFeat applied HOG as the prediction target instead of the RGB value. Other research endeavors \citep{el2021large,zhou2021ibot,assran2022masked, nips2021vatt,2022dilemma} combine the idea of CL with MIM. 
Moreover, Data2Vec~\citep{baevski2022data2vec} proposed a framework that applies the masked prediction idea for either speech, NLP, or CV.
However, most MIM works are confined to ViTs, recently proposed CIM \citep{fang2022corrupted} uses the output of a pre-trained tokenizer as the target and takes the output of a frozen BEiT as the encoder's input as a workaround to enable MIM on CNNs, and the concurrent work SparK~\citep{iclr2023spark} employs the sparse convolution operators to tackle the irregular masked input for CNNs.

\begin{figure*}[t]  
\centering
\vspace{-0.5em}
    \subfigtopskip=-0.5pt
    \subfigbottomskip=-0.5pt
    \subfigcapskip=-4pt
    \subfigure[]{\label{fig:in100_mask_a}\includegraphics[height=0.211\linewidth,trim= 5 2 0 0,clip]{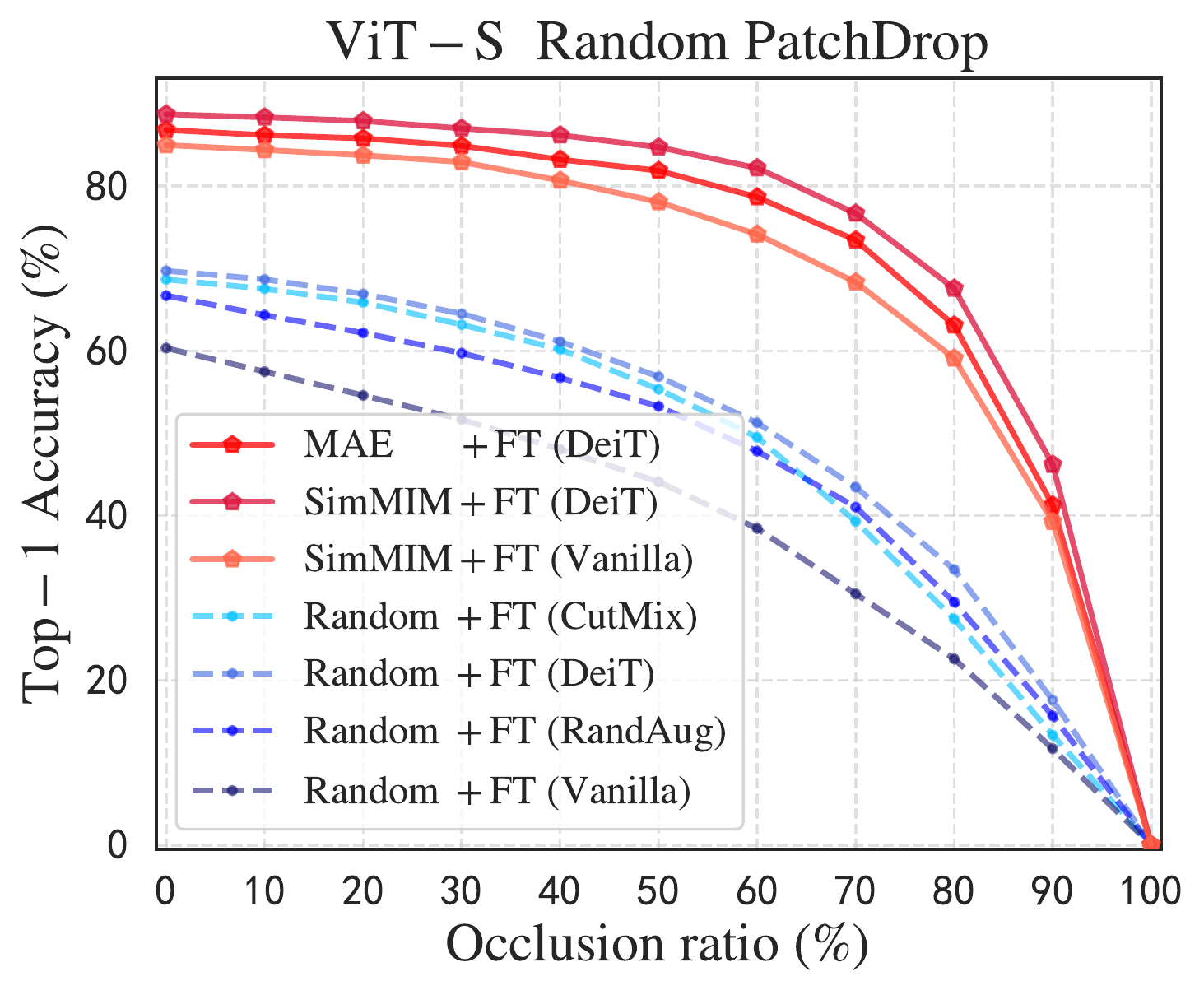}}
    \hspace{-0.25cm}
    \subfigure[]{\label{fig:in100_mask_c}\includegraphics[height=0.211\linewidth,trim= 5 2 0 0,clip]{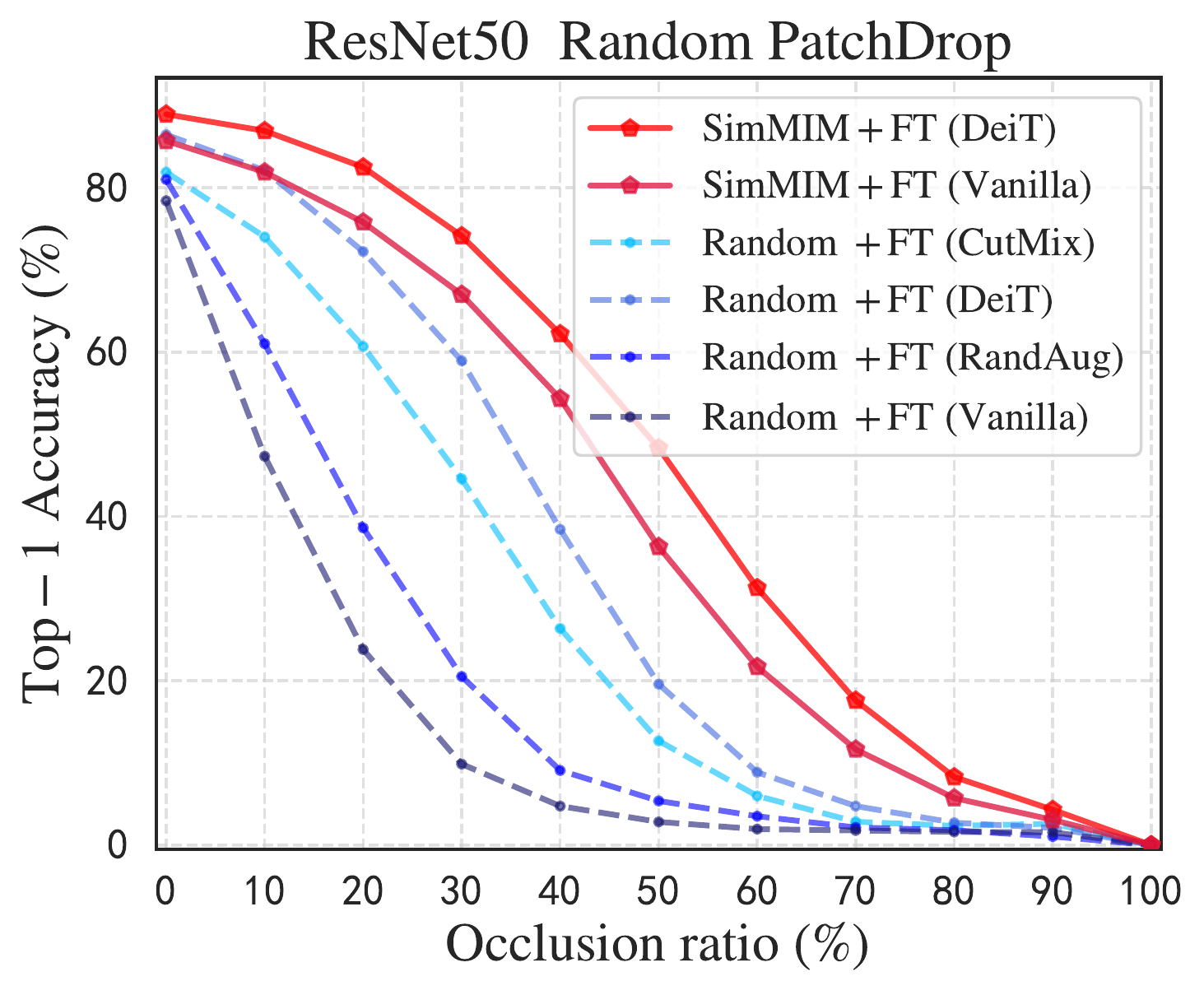}}
    \hspace{-0.25cm}
    \subfigure[]{\label{fig:in100_mask_e}\includegraphics[height=0.211\linewidth,trim= 5 2 0 0,clip]{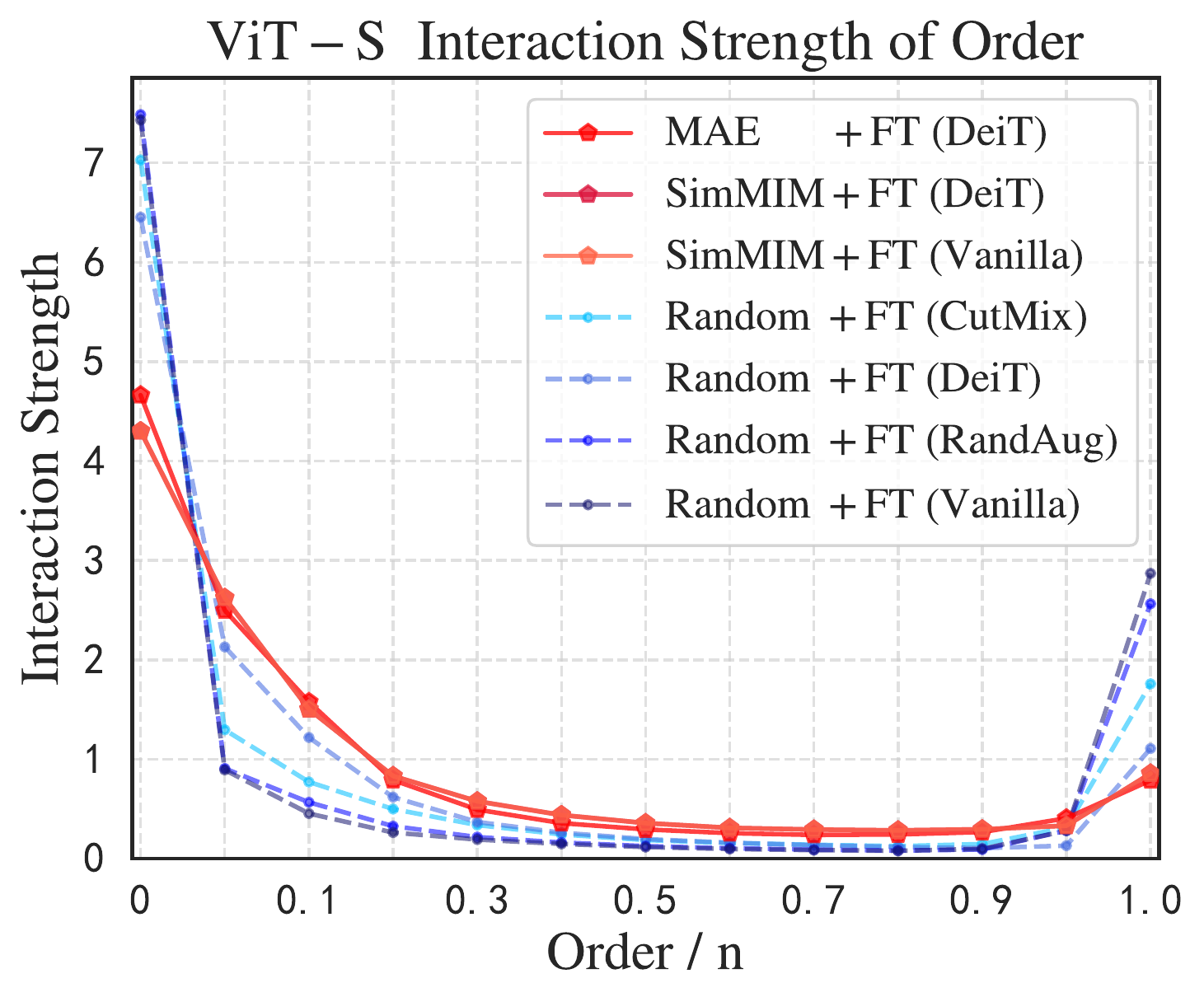}}
    \hspace{-0.25cm}
    \subfigure[]{\label{fig:in100_mask_g}\includegraphics[height=0.211\linewidth,trim= 5 2 0 0,clip]{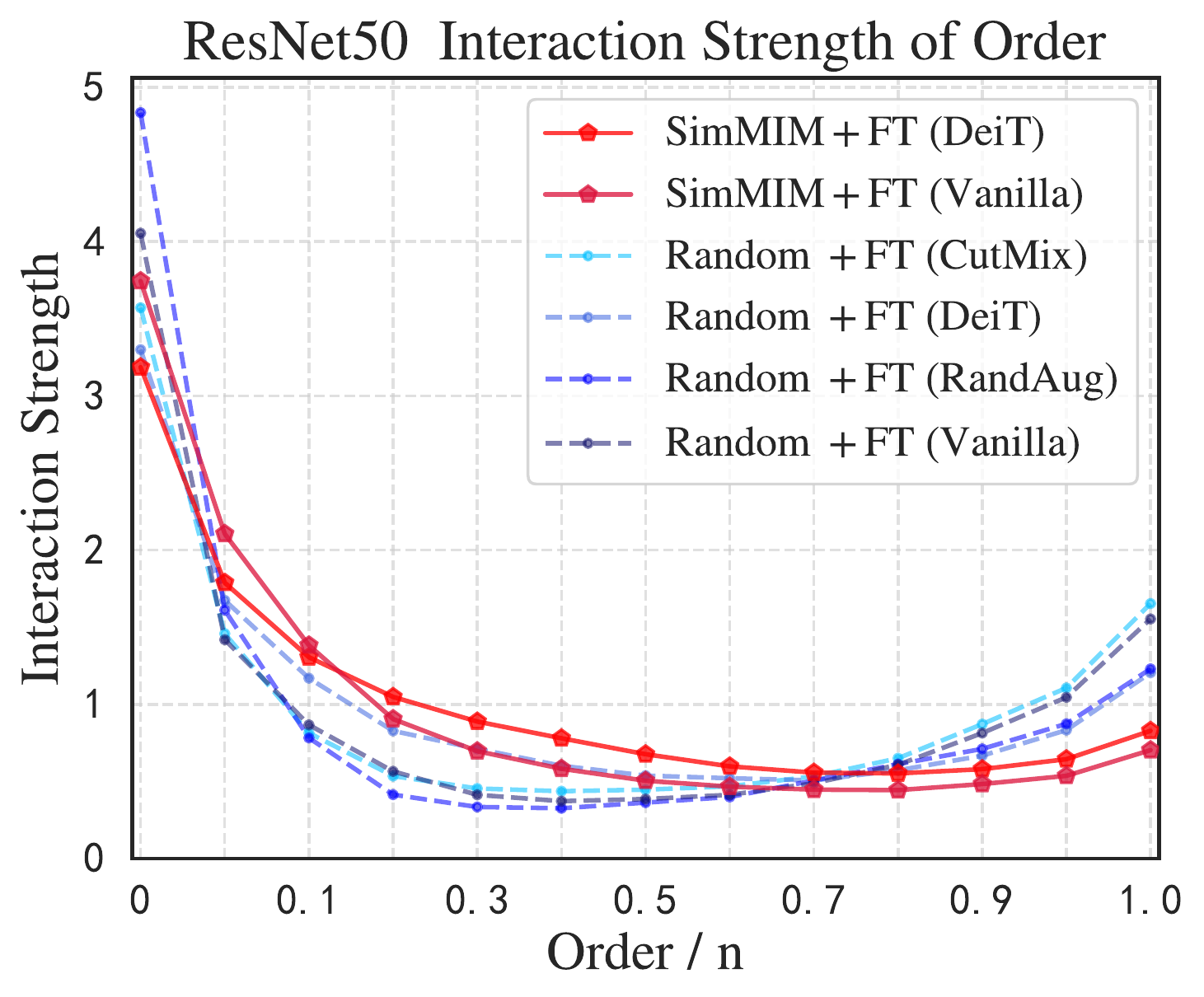}}
\vspace{-1.5em}
    \caption{(a)(b): Robustness against different occlusion ratios of images is studied for both ViT-S and ResNet-50 under different experimental settings (see Section \ref{sec:occlusion_robustness}). (c)(d): Distributions of the interaction strength $J^{(m)}$ are explored for both ViT-S and ResNet-50 under different experimental settings. The label indicates the pre-training method $+$ fine-tuning augmentation used, random stands for random weight initialization. Appendix~\ref{app:empirical_exp} provides more results and implement details.}
    \label{fig:in100_mask_interact}
    \vspace{-0.5em}
\end{figure*}

\vspace{-1.5em}
\section{Midst of Masked Image Modeling}
\label{sec:empirical}
\subsection{Is MIM Better Image Augmentation?}
\label{sec:occlusion_robustness}
Compared to CNN, Transformers gain tremendous performance improvement with carefully designed image augmentation techniques\citep{cubuk2020randaugment, yun2019cutmix, zhong2020random}. For instance, Random erasing and Cutmix randomly remove part of the image and replace the corresponding region with either Gaussian noise or a patch from another image. Similarly, as in most MIM pre-training tasks, some image patches are masked out and replaced with a learnable mask token. Noticing the resemblance of the masking operations, \textit{we hypothesize that MIM as a pre-training task and masking-based data augmentations enhance the network's robustness towards occlusion, enabling the network with a more generalized feature extraction ability.}
To verify our hypothesis, we design an occlusion robustness test. Let $x\in \mathbb{R}^{3\times H\times W}$ be an input image and $y\in \mathbb{R}^{C}$ be its corresponding label, where $C$ is the class number. Considering a classification task $y = f(x)$ where $f$ denotes a neural network, the network is considered robust if the network outputs the correct label given an occluded version of the image $x'$, namely $y = f(x')$. For occlusion, we consider the patch-based random masking as adopted in most MIM works \citep{he2021masked, xie2021simmim, wei2021masked}. In particular, we split the image of size $224\times224$ into patch size $16\times 16$ and randomly mask $M$ patches out of the total number of $N$ patches. The occlusion ratio could then be defined as $\frac{M}{N}$. We conduct experiments on ImageNet-100 (IN-100)~\citep{krizhevsky2012imagenet} for both Transformer and CNN with different settings. We choose ViT-S~\citep{dosovitskiy2020image} and ResNet-50\citep{he2016deep} as the network architecture. Robustness is compared under the following settings: \textbf{(i)} random weight initialization with no image augmentation applied; \textbf{(ii)} random weight initialization with different image augmentations applied; \textbf{(iii)} MIM pre-training as weight initialization with and without image augmentations applied.
In Fig.~\ref{fig:in100_mask_interact}, we report the average top-1 accuracy across five runs trained with different settings under various occlusion ratios. Fig.~\ref{fig:in100_mask_a} and \ref{fig:in100_mask_c} show that both MIM and patch-removing alike augmentations significantly improve model occlusion robustness for both ViT-S and ResNet-50. Nevertheless, MIM yields more robust feature extraction than adopting augmentations. Although MIM and patch-removing alike augmentations share similar masking mechanisms, MIM explicitly forces the model to learn the interactions between patches in order to reconstruct missing patches enabling more robust feature extraction. Comparing Fig.~\ref{fig:in100_mask_a} and \ref{fig:in100_mask_c}, the convex trend of accuracy from ViT-S indicates better robustness than the concave trend from ResNet-50. This can be attributed to the higher degrees of freedom of the self-attention mechanism compared to convolution priors. \textit{We claim that the success of MIM on ViTs can be seen as resonance in terms of better patch interactions imposed by MIM while supported by the self-attention mechanism of ViTs.}

\begin{figure*}[t]  
    \centering
    \includegraphics[width=0.97\linewidth]{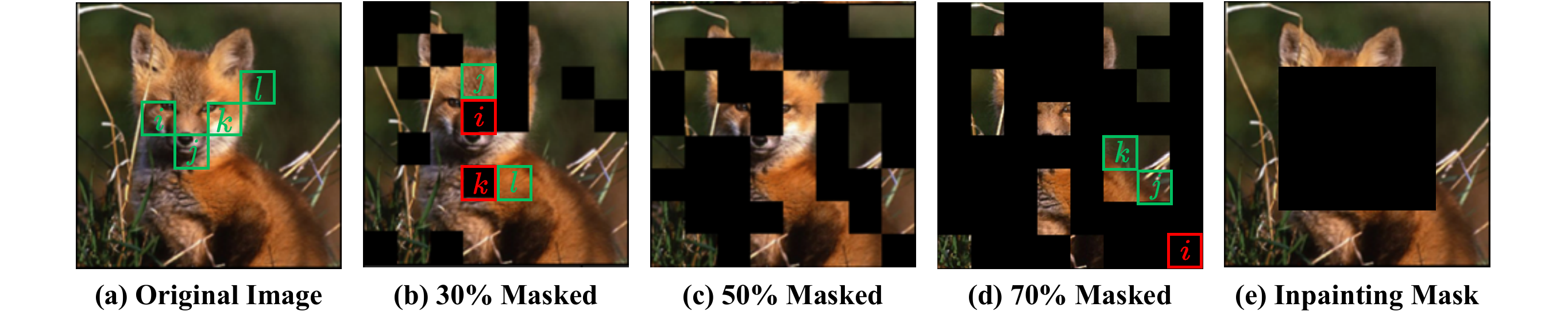}
    \vspace{-1.25em}
    \caption{(a) Four patches $(i,j,k,l)$ interact with each other and forms a contour or edge pattern of the fox for image categorization. (b) Image with 30\% masking ratio. Masked patches $i$ and $k$ interact with neighboring patches $j$ and $l$ to predict the missing patches. (c) Image with 50\% masking ratio. Masked patches force the model to extract information from unmasked patches and learn middle-order interactions for the MIM task. (d) Image with 70\% masking ratio. Masked Patch $i$ interacts with longer-range patches $j$ and $k$, forming an edge pattern. (e) A typical masking pattern for existing inpainting tasks.}
    \label{fig:masking_ratio}
    \vspace{-1.0em}
\end{figure*}

\subsection{Middle-order Interactions for Generalized Feature Extraction}
\label{sec:interaction}
Next, we show that MIM essentially enables better middle-order interactions between patches. Note that existing MIM works adopt a medium or high masking ratio~\citep{xie2021simmim,he2021masked} (\textit{e.g.}, 60\% or 70\%, see Fig.~\ref{fig:masking_ratio}) during pre-training, and in these settings, the pairwise interactions between patches are under a middle-size context measured by the order $m$. Early inpainting work based on CNN~\citep{pathak2016context} resembles MIM but attracts little attention due to limited performance. The inpainting task adopts the masking strategy as illustrated in Fig.~\ref{fig:in100_mask_e}, which masks a full large region instead of random small patches. Such masking mechanisms ignore patch interaction and focus only on reconstruction leading to poor representation quality. To investigate whether MIM makes the model more sensitive to patch interactions of some particular orders, we resort to the tool of multi-order interactions introduced by~\citep{deng2021discovering, zhang2020interpreting}. Intuitively, $m^{th}$-order interactions of patches refer to inference patterns (deep features) induced from $m$ number of patches of the original image in the input space. With a small value of $m$ (low-order interactions), the model simply learns local features such as texture.
Formally, the multi-order interaction $I^{(m)}(i,j)$ is to measure the order of interactions between patches $i$ and $j$. We define $I^{(m)}(i,j)$ to be the average interaction utility between patches $i$ and $j$ on all contexts consisting of $m$ patches, where $m$ denotes the order of contextual complexity of the interaction. 
Mathematically, given an input image $x$ with a set of $n$ patches $N = \{1,\dots,n\}$ (\textit{e.g.}, $n$ pixels), the multi-order interaction $I^{(m)}(i,j)$ is defined as:
\begin{equation}
\begin{aligned}
I^{(m)}(i,j) = \mathbb{E}_{S \subseteq N \setminus \{i,j\}, |S|=m}[\Delta f(i,j,S)],
\end{aligned}
\label{eq1}
\end{equation}
where $\Delta f(i,j,S) = f(S \cup \{i,j\}) - f(S \cup \{i\}) - f(S \cup \{j\}) + f(S)$. $f(S)$ indicates the score of output with patches in $N \setminus S$ kept unchanged but replaced with the baseline value~\citep{ancona2019explaining}, where the context $S\subseteq N$. See Appendix~\ref{app:interaction} for details. To measure the interaction complexity of the neural network, we measured the relative interaction strength $J^{(m)}$ of the encoded $m$-th order interaction as:
\vspace{-0.5em}
\begin{equation}
\begin{aligned}
J^{(m)} = \frac{\mathbb{E}_{x \in \Omega}\mathbb{E}_{i,j}|I^{(m)}(i,j|x)|}{\mathbb{E}_{m^{'}}\mathbb{E}_{x \in \Omega}\mathbb{E}_{i,j}|I^{(m^{'})}(i,j|x)|},
\vspace{-0.5em}
\end{aligned}
\label{eq2}
\end{equation}
where $\Omega$ is the set of all samples and $0\le m \ge n-2$. $J^{(m)}$ is the average value over all possible pairs of patches of input samples. $J^{(m)}$ is normalized by the average value of all interaction strengths. $J^{(m)}$ then indicates the distribution (area under curve sums up to one) of the order of interactions of the network. We use $J^{(m)}$ as the metric to evaluate and analyze interaction orders of the network with MIM pre-training. We conduct experiments on IN-100 with image size $224\times 224$ and use ViT-S~\citep{dosovitskiy2020image} and ResNet-50~\citep{he2016deep} as the network architecture. We consider a patch of size $16\times 16$ as input. For the computation of $J^{(m)}$, we adopt the sampling solution following previous works~\citep{deng2021discovering, zhang2020interpreting}. As can be seen from Fig.~\ref{fig:in100_mask_e}, ViT-S with random weight initialization tends to learn simple interactions with few patches (e.g., less than $0.05n$ patches) while MIM pre-trained models show a stronger interaction for relative middle-order (from $0.05n$ to $0.5n$). Similarly, as observed from \ref{fig:in100_mask_g}, MIM pre-trained ResNet-50 enhances the middle-order interactions from $0.1n$ to $0.55n$ compared to random initialized models.
Stronger middle-order interactions form more complex features such as shape and edge compared to local texture features learned from low-order interactions~\citep{naseer2021intriguing}.

\section{Approach}
\label{sec:method}
\begin{figure*}[t!]
    \vspace{-0.5em}
\centering
    \includegraphics[width=\linewidth]{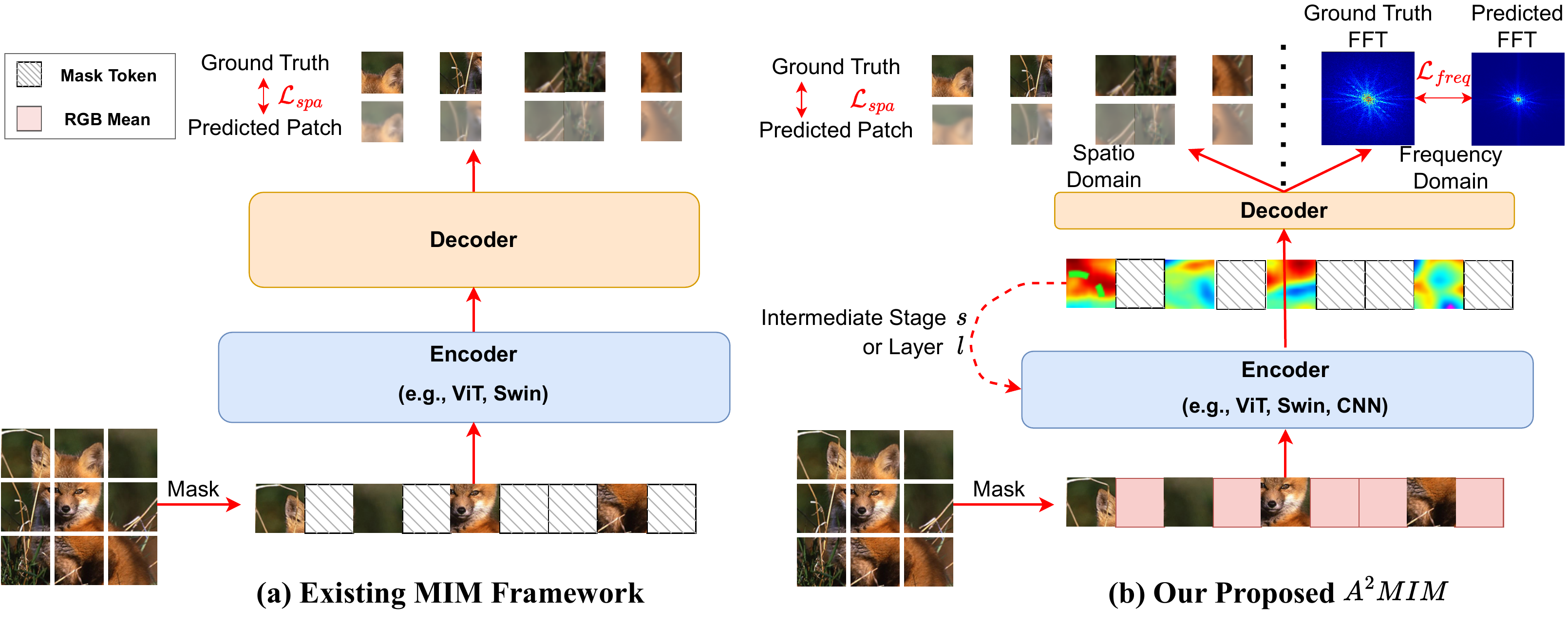}
    \vspace{-2.75em}
    \caption{
    An illustration comparison between the existing MIM framework and our proposed framework. For the existing MIM framework, the input image is patchfied into a sequence of patches without overlapping with masked patches that are replaced with learnable mask tokens. The sequence is then input to the Transformer encoder. The $\mathcal{L}_{spa}$ is applied between the ground truth patches and the reconstructed patches from the decoder in the spatiotemporal domain. Our proposed framework uses the mean RGB value of the image instead of the mask token in the input space. We then add a learnable mask token onto the intermediate feature map of layer-$l$ of stage-$s$ of the encoder instead of replacement in the input space. The encoder could either be of the Transformer or the CNN family. In addition to the $\mathcal{L}_{spa}$, we adopt a $\mathcal{L}_{freq}$ in the Fourier domain to enhance the encoder to learn more middle-order interactions. Specifically, we apply DFT on both the ground truth image and the predicted image and then use Mean square error (MSE) to measure the difference.}
    \vspace{-0.75em}
    \label{fig:framework}
\end{figure*}

We propose a generic MIM framework following two design rules: (a) \textbf{Better middle-order interactions between patches for more generalized feature extraction.} (b) \textbf{No complex or non-generic designs are adopted to ensure compatibility with all network architectures.} Figure~\ref{fig:framework} highlights the difference between A$^2$MIM and existing MIM frameworks in terms of three key components: masking strategy, encoder/decoder network architecture design and prediction targets.

\subsection{Architecture Agnostic Framework}
\paragraph{Mask Where Middle-order Interactions Occur.}
Existing works~\citep{el2021large, he2021masked, wei2021masked} adopt the masking strategy where the input image is divided into non-overlapping patches, and a random subset of patches is masked. MAE utilizes a Transformer as a decoder and takes only the visible patches into the encoder. Masked tokens are appended to the decoder to reconstruct the masked patches. SimMIM and MaskFeat~\citep{wei2021masked} utilize a fully connected layer as the decoder and feed the mask token into the encoder together with the visible patches. The mask token~\citep{devlin2018bert} is a token-shared learnable parameter that indicates the presence of missing patches to be predicted. Despite different choices of decoder structures, the mask token is either placed at the input to the encoder or the decoder. Mathematically, the masking process of MIM is defined as $x_{mask} = x \odot (1 - M) + T \odot M$, where $M$ is the random occlusion mask, and $T$ represents the learnable mask token. Such masking at the patch embedding layer aligns with the attention mechanism of Transformers, which is robust against occlusion. However, masking at the stem layer undermines the context extraction capability of CNN, which relies on local inductive biases. Moreover, masking at input stages of the network leads to low-order interactions.
Thus, we propose to mask intermediate features where the output feature contains both semantic and spatial information, and the mask token can encode interactions with a medium number of tokens (\textit{e.g.}, the last embedded stage).
Concretely, our masking operation is defined as $z^{l}_{mask} = z^{l} + T \odot D(M)$, where $z^{l}$ is the intermediate feature map at stage-$l$ in CNN encoders (or layer-$l$ in Transformers) and $D(\cdot)$ is the corresponding down-sampling function of the occlusion mask.

\vspace{-0.5em}
\paragraph{Filling Masked Tokens with RGB Mean.}
Existing works directly replace the occluded patches with the mask token in the input space or after the patch embedding~\citep{bao2021beit, xie2021simmim}. In contrast, we use the average RGB value to fill the occluded patches as the input to the encoder and add the mask token onto the intermediate feature maps of the encoder. The masking mechanism originates from NLP where languages are of high-level semantics and do not require low-level feature extraction as image processing. Masking at the early stages of the network where low-level feature extraction happens is harmful in terms of feature extraction.
The RGB mean is the DC component of images. Filling with RGB mean alleviates local statistics distortion caused by the masking operation and forces the network to model more informative medium frequencies instead of filling the masked patches with blurry color blocks (low frequencies). The proposed masking strategy is generic to both convolution and self-attention in that it accommodates low-level to semantic-level feature extraction.

\subsection{Middle-order Interactions from Fourier Perspective}
Current works~\citep{el2021large, he2021masked, xie2021simmim} adopt raw RGB values as the prediction target. However, raw pixels in the spatial domain are heavily redundant and often contain low-order statistics~\citep{bao2021beit, wei2021masked, zhou2021ibot}. MaskFeat~\citep{wei2021masked} adopts the Histogram of Oriented Gradients (HOG) as the prediction target outperforming MAE and SimMIM. HOG is a discrete descriptor of medium or high-frequency features that captures shape patterns based on middle-order interactions.
ViTs and CNNs have low-pass and high-pass filtering properties, respectively \citep{park2022vision, park2022blur}. ViTs and CNNs have certain frequency bands that they each cannot model well, and both cannot model middle-order interactions well (detailed in Appendix \ref{app:analysis_freq}). The observation of the medium frequency descriptor HOG improves middle-order interactions and leads to the hypothesis that learning medium frequencies would help the model learn more middle-order interactions.
Given a RGB image $x \in \mathbb{R}^{3\times H\times W}$, the discrete Fourier transform (DFT) of each channel is defined as:
\begin{equation}
\begin{aligned}
F_{(u,v)} = \sum_{h=1}^{h=H}\sum_{w=1}^{w=W} 
x(h,w) e^{-2\pi j (\frac{uh}{H} + \frac{vw}{W})}.
\end{aligned}
\label{eq:DFT}
\end{equation}
In addition to the common MIM loss in the spatial domain $\mathcal{L}_{spa}$,
we propose $\mathcal{L}_{freq}$ in Fourier domain:
\begin{equation}
\begin{aligned}
\mathcal{L}_{freq} = \sum_{c=1}^{c=3}\sum_{u=1}^{u=H}\sum_{w=1}^{w=W} \omega(u,v) \big\lVert \mathrm{DFT}(x^{pred}_{c} \odot M +\\
    \mathrm{de} (x^{pred}_{c}) \odot (1 - M)) - \mathrm{DFT}(x_{c})\big\lVert,
\end{aligned}
\label{eq:fft_loss}
\end{equation}
where $x^{pred}$ is the predicted image, $\mathrm{de}(\cdot)$ is \texttt{detach} gradient operation, and $\omega(u,v)$ is a dynamic frequency weighting matrix. Inspired by Focal Frequency loss~\citep{jiang2021focal}, we define adaptive $\omega(u,v)$ as follows:
\begin{equation}
\begin{aligned}
\omega(u,v) = &\big\lVert \mathrm{DFT}\big(x^{pred}_{c} \odot M +\\
            &\det(x^{pred}_{c}) \odot (1 - M)\big) - \mathrm{DFT}(x_{c}) \big\lVert,
\end{aligned}
\label{eq:reweight}
\end{equation}
$\omega(u,v)$ enables both ViTs and CNNs to model features of medium frequencies rather than local textures and noise corresponding to high frequencies. Since filling masked tokens with RGB mean is filling with DC components, combining our proposed masking strategy with the weighting effect of the $\mathcal{L}_{freq}$ leads to the better modeling of medium frequency features (middle-order interactions). Fig.~\ref{app:analysis_freq} verifies that Eq.~(\ref{eq:reweight}) allows the model to learn previously ignored frequencies (mostly the medium frequencies).
Note that $\mathcal{L}_{freq}$ introduces negligible overhead by using Fast Fourier Transform (FFT) algorithms with $\mathcal{O} (n\log n)$ complexity to achieve DFT.
The overall loss of A$^2$MIM is defined as:
\begin{equation}
\begin{aligned}
\mathcal{L} = \mathcal{L}_{spa} + \lambda \mathcal{L}_{freq},
\end{aligned}
\label{eq:final}
\end{equation}
where $\mathcal{L}_{spa}$ = $\norm{x^{pred} - x} \odot M$ and $\lambda$ is a loss weighting hyper-parameter. We set $\lambda$ to 0.1 by default.

\section{Experiments}
\label{sec:exp}
\subsection{Pre-training Setup}
We adopt Vision Transformer~\citep{dosovitskiy2020image} (ViT/16), ResNet~\citep{he2016deep}, and ConvNeXt~\citep{cvpr2022convnext} as the backbone encoder. Models are pre-trained on ImageNet-1K (IN-1K) training set with AdamW~\citep{iclr2019AdamW} optimizer, a batch size of 2048, and a basic learning rate of $1.2\times 10^{-3}$ adjusted by a cosine learning rate scheduler. The input image size is $224\times 224$ with a masked patch size of $32\times 32$, and the random masking ratio is 60\%. By default, the learnable mask tokens are placed at stage-3 and layer-0 in ResNet/ConvNeXt and ViT architectures, respectively. We adopt a linear prediction head as the MIM decoder \citep{xie2021simmim}. A$^2$MIM+ indicates adopting HOG as the MIM target and using the MLP decoder with depth-wise (DW) convolutions. Our experiments are implemented on OpenMixup~\citep{li2022openmixup} by Pytorch and conducted on workstations with NVIDIA A100 GPUs. \textbf{Bold} and \ul{underline} indicate the best and the second-best performance, and \gray{gray} denotes the uncomparable results (\textit{e.g.}, not in the same technical scope). See Appendix~\ref{app:comparison_exp} for pre-training details.

\begin{table}[t!]
    \vspace{-0.5em}
    \setlength{\tabcolsep}{0.8mm}
    \centering
    \caption{ImageNet-1K fine-tuning (FT) top-1 accuracy (\%) of ViT-S and ViT-B models. $^\ddagger$ denotes our finetuned results.}
    \vspace{0.25em}
    \label{tab:in1k_vit_cls}
\resizebox{\linewidth}{!}{
    \begin{tabular}{l|lcc|ccc}
    \toprule
    Method                           & Date               & Target       & PT         & ViT-S       & ViT-B       & ViT-L       \\
                                     &                    &              & Epochs     & FT          & FT          & FT          \\ \hline
    \gray{Rand init.}                & -                  & \gray{Label} & \gray{300} & \gray{79.9} & \gray{81.8} & \gray{82.6} \\
    SimCLR                           & \small{ICML'2020}  & CL           & 300        & 80.2        & 82.3        & -           \\
    BYOL                             & \small{NIPS'2020}  & CL           & 300        & 80.9        & 82.8        & -           \\
    MoCoV3                           & \small{ICCV'2021}  & CL           & 300        & 81.4        & 83.2        & 84.1        \\
    DINO                             & \small{ICCV'2021}  & CL           & 300        & 81.5        & 83.6        & -           \\ \hline
    BEiT                             & \small{ICLR'2022}  & DALLE        & 800        & 81.3        & 83.2        & 85.2        \\
    SplitMask                        & \small{arXiv'2022} & DALLE        & 300        & 81.5        & 83.6        & -           \\
    iBOT                             & \small{ICLR'2022}  & EMA          & 800        & \bf{82.3}   & 84.0        & 85.2        \\
    MAE                              & \small{CVPR'2022}  & RGB          & 1600       & 81.6        & 83.6        & 85.9        \\
    MaskFeat                         & \small{CVPR'2022}  & HOG          & 800        & -           & 84.0        & 85.7        \\
    Data2Vec                         & \small{ICML'2022}  & EMA          & 800        & -           & 84.2        & 86.2        \\
    SimMIM                           & \small{CVPR'2022}  & RGB          & 800        & 81.7        & 83.8        & 85.6        \\
    CAE                              & \small{arXiv'2022} & DALLE        & 1600       & 81.8        & 83.6        & \ul{86.3}   \\
    $mc$-BEiT                        & \small{ECCV'2022}  & VQGAN        & 800        & -           & 84.1        & 85.6        \\
    BootMAE                          & \small{ECCV'2022}  & EMA          & 800        & -           & 84.2        & 85.9        \\
    PeCo                             & \small{AAAI'2023}  & VQVAE        & 800        & -           & \bf{84.5}   & \bf{86.5}   \\
    CIM                              & \small{ICLR'2023}  & BEiT         & 300        & 81.6        & 83.3        & -           \\
    MC-MAE                           & \small{ICLR'2023}  & EMA          & 1600       & 82.0        & 83.6        & 86.1        \\
    MAGE-C                           & \small{CVPR'2023}  & VQGAN        & 1600       & -           & 82.9        & 84.3        \\
    LocalMIM                         & \small{CVPR'2023}  & HOG          & 1600       & -           & 84.0        & 85.8        \\
    \rowcolor{gray94} \bf{A$^2$MIM}  & \bf{Ours}          & RGB          & 800        & \ul{82.1}   & 84.2        & 86.1        \\
    \rowcolor{gray94} \bf{A$^2$MIM+} & \bf{Ours}          & HOG          & 800        & \bf{82.3}   & \ul{84.4}   & \ul{86.3}   \\
    \bottomrule
    \end{tabular}
    }
    \vspace{-1.0em}
\end{table}

\begin{table}[t!]
    \vspace{-0.5em}
    \setlength{\tabcolsep}{0.5mm}
    \centering
    \caption{ImageNet-1K linear probing (Lin.) and fine-tuning (FT) top-1 accuracy (\%) of ResNet-50.
    \protect\\{\fontsize{8pt}{8pt}\selectfont $^\dag$Multi-crop augmentation. \quad $^\ddagger$Our modified MIM methods for CNN.}
    }
    \vspace{0.25em}
    \label{tab:in1k_cnn_cls}
\resizebox{\linewidth}{!}{
    \begin{tabular}{l|ccc|ccc}
    \toprule
    Method                           & \multicolumn{3}{c|}{Fast Pre-training} & \multicolumn{3}{c}{Longer Pre-training} \\
                                     & Epochs     & Lin.        & FT (A3)     & Epochs      & FT (A3)     & FT (A2)     \\ \hline
    \gray{Rand init.}                & -          & \gray{4.4}  & \gray{78.1} & -           & \gray{78.1} & \gray{79.8} \\
    \gray{PyTorch (Sup.)}            & \gray{90}  & \gray{76.2} & \gray{78.8} & \gray{300}  & \gray{78.9} & \gray{79.9} \\
    Inpainting                       & 70         & 40.1        & 78.4        & 300         & 78.0        & -           \\
    Relative-Loc                     & 70         & 38.8        & 77.8        & 300         & 77.9        & -           \\
    Rotation                         & 70         & 48.1        & 77.7        & 300         & 78.2        & -           \\ \hline
    SimCLR                           & 100        & 64.4        & 78.5        & 800         & 78.8        & 79.9        \\
    MoCoV2                           & 100        & 66.8        & 78.5        & 800         & 78.8        & 79.8        \\
    BYOL                             & 100        & 68.4        & 78.7        & 400         & \ul{78.9}   & 80.1        \\
    SwAV$^\dag$                      & 100        & 71.9        & \bf{78.9}   & 400         & \bf{79.0}   & 80.2        \\
    Barlow Twins                     & 100        & 67.2        & 78.5        & 300         & 78.8        & 79.9        \\
    MoCoV3                           & 100        & 68.9        & 78.7        & 300         & \bf{79.0}   & 80.1        \\ \hline
    BEiT$^\ddagger$                  & 100        & \gray{47.1} & 78.1        & -           & -           & -           \\
    Data2Vec$^\ddagger$              & 100        & \gray{43.2} & 78.0        & -           & -           & -           \\
    MAE$^\ddagger$                   & 100        & \gray{37.8} & 77.1        & 300         & 77.2        & 79.0        \\
    SimMIM$^\ddagger$                & 100        & \gray{47.5} & 78.2        & 300         & 78.3        & 79.9        \\
    CIM                              & -          & -           & -           & 300         & 78.6        & \ul{80.4}   \\
    \rowcolor{gray94} \bf{A$^2$MIM}  & 100        & \gray{48.1} & \ul{78.8}   & 300         & \ul{78.9}   & \ul{80.4}   \\
    \rowcolor{gray94} \bf{A$^2$MIM+} & 100        & \gray{50.3} & \bf{78.9}   & 300         & \bf{79.0}   & \bf{80.5}   \\
    \bottomrule
    \end{tabular}
    }
    \vspace{-1.0em}
\end{table}

\subsection{Image Classification on ImageNet-1K}
\label{sec:exp_in1k}
\paragraph{Evaluation Protocols.}
We evaluate the learned representation by end-to-end fine-tuning (FT) and linear probing (Lin.) protocols on IN-1K. For FT evaluations of ViTs, 
we employ the fine-tuning as MAE~\citep{he2021masked}, which applies DeiT~\citep{touvron2021training} augmentations, AdamW optimizer with a batch size of 1024 for 200 epochs, and adopt a layer-wise learning rate decay of 0.65 as BEiT~\citep{bao2021beit}. For FT evaluations of CNNs, ResNet variants are fine-tuned with RSB A2/A3~\citep{wightman2021rsb} training settings, which employ LAMB~\citep{iclr2020lamb} optimizer with a batch size 2048 for 300/100 epochs, and ConvNeXt models are fine-tuned 300-epoch with its original supervised learning settings. For the linear evaluations, ResNet-50 settings follow MoCo~\citep{cvpr2020moco}, which trains a linear classifier by SGD with a batch size of 256, and ViTs follow MAE, which tunes the linear layer with BN by AdamW.
See Appendix~\ref{app:comparison_exp} for detailed configurations.

\begin{table}[t]
    \vspace{-0.5em}
    \setlength{\tabcolsep}{0.7mm}
    \centering
    \caption{ImageNet-1K fine-tuning (FT) top-1 accuracy (\%) with ResNet and ConvNeXt of various model scales. We adopt the 300-epoch fine-tuning protocols for both architectures. $^\ddagger$ denotes our reproduced results.}
    \label{tab:in1k_cnn_extend}
\resizebox{1.0\linewidth}{!}{
    \begin{tabular}{l|c|cccc >{\columncolor{gray94}}c}
    \toprule
    Methods     & \#Para. & \gray{Sup.}  & MoCoV3$^\ddagger$ & SimMIM$^\ddagger$ & \gray{SparK} & A$^2$MIM \\
    Target      & (M)     & \gray{Label} & CL                & RGB               & \gray{RGB}   & RGB        \\ \hline
    ResNet-50   & 25.6    & \gray{79.8}  & 80.1              & 79.9              & \gray{80.6}  & \bf{80.4}  \\
    ResNet-101  & 44.5    & \gray{81.3}  & 81.6              & 81.3              & \gray{82.2}  & \bf{81.9}  \\
    ResNet-152  & 60.2    & \gray{81.8}  & 82.0              & 81.9              & \gray{82.7}  & \bf{82.5}  \\
    ResNet-200  & 64.7    & \gray{82.1}  & 82.5              & 82.2              & \gray{83.1}  & \bf{83.0}  \\ \hline
    ConvNeXt-T  & 28.6    & \gray{82.1}  & 82.3              & 82.1              & \gray{82.7}  & \bf{82.5}  \\
    ConvNeXt-S  & 50.2    & \gray{83.1}  & 83.3              & 83.2              & \gray{84.1}  & \bf{83.7}  \\
    ConvNeXt-B  & 88.6    & \gray{83.5}  & 83.7              & 83.6              & \gray{84.8}  & \bf{84.1}  \\
    \bottomrule
    \end{tabular}
    }
    \vspace{-1.0em}
\end{table}


\vspace{-0.75em}
\paragraph{ViTs.}
We first evaluate A$^2$MIM variants with ViT-S/B/L on IN-1K. We list the supervision target used by various pre-training algorithms in the third column of Tab.~\ref{tab:in1k_vit_cls}. VQVAE/DALL-E~\citep{2021dalle} and VQGAN~\citep{cvpr2021vqgan} are pre-trained image tokenizers, while EMA refers to the momentum encoder. Our A$^2$MIM outperforms CL and MIM baselines, and A$^2$MIM+ achieves competitive results as current state-of-the-art methods with complex supervision, \textit{e.g.,} SplitMask (MIM with CL combined), iBOT (complex teacher-student architecture), and CIM (pre-trained BEiT as supervision). Based on ViT-S/B/L, A$^2$MIM significantly improves the baseline SimMIM by 0.5\%/0.4\%/0.5\% with the RGB target and 0.7\%/0.7\%/0.6\% with the HOG feature as supervision.

\vspace{-0.75em}
\paragraph{CNNs.}
We then compare A$^2$MIM with classical self-supervised learning methods (Inpainting~\citep{pathak2016context}, Relative-Loc~\citep{iccv2015relativeloc}, and Rotation~\citep{iclr2018rotation}), CL, and MIM methods with 100/300 pre-training epochs. We modified MIM methods to run them on ResNet-50: the learnable mask token is employed to the encoder for BEiT~\citep{bao2021beit}, Data2Vec~\citep{baevski2022data2vec}, and SimMIM~\citep{xie2021simmim} after the stem (the output feature of $56\times 56$ resolutions); the encoder of MAE randomly selects 25\% from $56\times 56$ output features of the stem as unmasked patches and takes the reorganized $28\times 28$ patches as the input of four stages. In Tab.~\ref{tab:in1k_cnn_cls}, our approach achieves competitive performance with state-of-the-art contrastive-based methods under 100-epoch FT evaluation. Note that MIM methods see fewer training samples per epoch than CL methods (\textit{e.g.}, 40\% \textit{vs.} 200\% of patches) and usually require longer pre-training epochs. Based on a longer FT evaluation, A$^2$MIM (300-epoch) outperforms contrastive-based methods with even fewer training epochs. Meanwhile, A$^2$MIM also improves the baseline SimMIM$^{\dagger}$ (+0.8\%) and the concurrent work CIM (+0.4\%) in terms of 100-epoch FT for the longer pre-training. Besides, we also report the linear probing (Lin.) results of the fast pre-training for reference, although we focus on learning representations with better fine-tuning performances. Although A$^2$MIM achieves lower Lin. results than popular CL methods, A$^2$MIM still improves the baseline by 0.6\%.
Moreover, we further conduct scaling-up experiments of A$^2$MIM and pre-training methods based on ResNet and ConvNeXt models. Notice that two concurrent works proposed after our A$^2$MIM (SparK~\citep{iclr2023spark} and ConvNeXtV2~\citep{Woo2023ConvNeXtV2}) are specially designed MIM approaches for CNNs, which employ the sparse convolution to handle the irregular masked input. As shown in Table~\ref{tab:in1k_cnn_extend}, we compare A$^2$MIM with DeiT (as the supervised baseline), MoCoV3, SimMIM, and SparK, where A$^2$MIM noticeably surpasses the two popular self-supervised methods (MoCoV3 and SimMIM). Despite the proposed A$^2$MIM yields inferior performances than SparK, A$^2$MIM can also work for Transformer architectures.

\begin{table}[t]
    \vspace{-0.5em}
    \setlength{\tabcolsep}{0.8mm}
    \centering
    \caption{Performance of object detection and semantic segmentation tasks based on ViT-B on COCO and ADE-20K.}
    \label{tab:downstream_vit}
    \vspace{1pt}
\resizebox{\linewidth}{!}{
    \begin{tabular}{l|cc|cccc}
    \toprule
    Method                          & Target       & Epochs     & IN-1K       & \multicolumn{2}{c}{COCO}  & ADE-20K     \\
                                    &              & PT         & FT          & AP$^{box}$  & AP$^{mask}$ & mIoU        \\ \hline
    \gray{DeiT (Sup.)}              & \gray{Label} & \gray{300} & \gray{81.8} & \gray{47.9} & \gray{42.9} & \gray{47.0} \\
    MoCoV3                          & CL           & 300        & 83.2        & 47.9        & 42.7        & 47.3        \\
    DINO                            & CL           & 400        & 83.6        & 46.8        & 41.5        & 47.2        \\ \hline
    BEiT                            & DALLE        & 300        & 83.2        & 43.1        & 38.2        & 47.1        \\
    iBOT                            & EMA          & 400        & 84.0        & 48.4        & 42.7        & 48.0        \\
    PeCo                            & VQ-VAE       & 300        & \bf{84.5}   & 43.9        & 39.8        & 46.7        \\
    MAE                             & RGB          & 1600       & 83.6        & 48.5        & 42.8        & 48.1        \\
    MaskFeat                        & HOG          & 800        & 84.0        & \ul{49.2}   & 43.2        & \ul{48.8}   \\
    SimMIM                          & RGB          & 800        & 83.8        & 48.9        & 43.0        & 48.4        \\
    CAE                             & DALLE        & 800        & 83.6        & \ul{49.2}   & \ul{43.3}   & \ul{48.8}   \\
    \rowcolor{gray94} \bf{A$^2$MIM} & RGB          & 800        & \ul{84.2}   & \bf{49.4}   & \bf{43.5}   & \bf{49.0}   \\
    \bottomrule
    \end{tabular}
    }
    \vspace{-1.5em}
\end{table}

\begin{table}[b]
    \vspace{-2.0em}
    \setlength{\tabcolsep}{1.1mm}
    \centering
    \caption{Performance of object detection and semantic segmentation tasks based on ResNet-50 on COCO and ADE20K.}
    \vspace{1pt}
    \label{tab:downstream_cnn}
\resizebox{\linewidth}{!}{
    \begin{tabular}{l|cc|cccc}
    \toprule
    Method                          & Target       & Epochs     & IN-1K       & \multicolumn{2}{c}{COCO}  & ADE-20K     \\
                                    &              & PT         & FT          & AP$^{box}$  & AP$^{mask}$ & mIoU        \\ \hline
    \gray{Sup.}                     & \gray{Label} & \gray{90}  & \gray{79.8} & \gray{38.2} & \gray{33.3} & \gray{36.1} \\
    SimCLR                          & CL           & 800        & 79.9        & 37.9        & 33.3        & 37.6        \\
    MoCoV2                          & CL           & 800        & 79.8        & \ul{39.2}   & 34.3        & 37.5        \\
    BYOL                            & CL           & 400        & 80.1        & 38.9        & 34.2        & 37.2        \\
    SwAV                            & CL           & 800        & \ul{80.2}   & 38.4        & 33.8        & 37.3        \\
    SimSiam                         & CL           & 400        & 80.0        & \ul{39.2}   & \ul{34.4}   & 37.2        \\
    Balow Twins                     & CL           & 800        & 79.9        & \ul{39.2}   & 34.3        & 37.3        \\ \hline
    SimMIM$^\ddagger$               & RGB          & 300        & 79.9        & 39.1        & 34.2        & 37.4        \\
    CIM                             & BEiT         & 300        & \bf{80.4}   & -           & -           & \ul{38.0}   \\
    \rowcolor{gray94} \bf{A$^2$MIM} & RGB          & 300        & \bf{80.4}   & \bf{39.8}   & \bf{34.9}   & \bf{38.3}   \\
    \bottomrule
    \end{tabular}
    }
\end{table}

\begin{figure*}[t!]
    \vspace{-0.5em}
    \centering
    \includegraphics[width=1.0\linewidth]{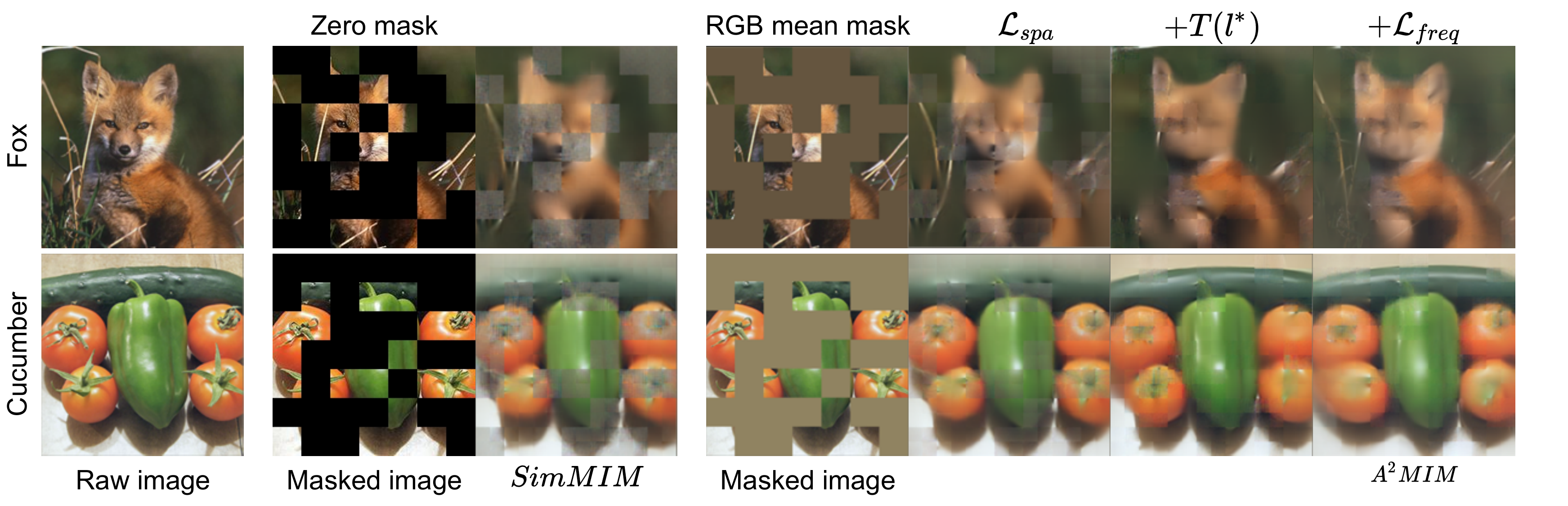}
    \vspace{-2.5em}
    \caption{
    Visualizations of predicted results from SimMIM (middle) and our A$^2$MIM (right) based on ViT-S pre-trained 400-epochs on IN-1K. Notice that $T(l^*)$ denotes the mask token $T$ to the optimal layer-5 in ViT-S. We ablate the proposed components by adding them to the baseline. Compared to results from SimMIM, reconstruction results of the modified baseline ($\mathcal{L}_{spa}$) with the RGB mean mask relieves grid-like artifacts; adding the mask token $T(l^*)$ further improves the smoothness; using the proposed $\mathcal{L}_{freq}$ helps the model to capture more informative details and contours.}
    \label{fig:ablation_rec_vit}
    \vspace{-1.0em}
\end{figure*}

\subsection{Transfer Learning Experiments}
\label{sec:exp_transfer}
\paragraph{Object detection and segmentation on COCO.}
To verify the transferring abilities, we benchmark CL and MIM methods on object detection and segmentation with COCO~\citep{eccv2014MSCOCO}. For evaluation on CNN, we follow the setup in MoCo, which fine-tunes Mask R-CNN~\citep{2017iccvmaskrcnn} with ResNet-50-C4 backbone using 2$\times$ schedule on the COCO~\textit{train2017} and evaluates on the COCO~\textit{val2017}. Results in Tab.~\ref{tab:downstream_cnn} indicate that A$^2$MIM (300-epoch) outperforms contrastive-based methods with longer pre-training (+0.7\% AP$^{box}$ and +0.6\% AP$^{mask}$). For evaluation on Transformer, we follow MAE and CAE, which efficiently fine-tunes Mask R-CNN with ViT-B backbone using 1$\times$ schedule. In Tab.~\ref{tab:downstream_vit}, A$^2$MIM (800-epoch) is superior to popular contrastive-based and MIM methods, \textit{e.g.}, outperforms MAE (1600-epoch) by 0.9\% AP$^{box}$ and 0.8\% AP$^{mask}$. 

\vspace{-0.75em}
\paragraph{Semantic segmentation on ADE20K.}
We then evaluate the transferring performances on semantic segmentation with ADE20K~\citep{ijcv2019ADE20K} by fine-tuning FCN~\citep{tpami2017fcn} and UperNet~\citep{eccv2018upernet}. Based on ResNet-50, all models are fine-tuned for 160K iterations with SGD following MoCo and CIM. Results in Tab.~\ref{tab:downstream_cnn} show that our method outperforms CL methods by at least 0.9\% mIoU and outperforms CIM (required extra pre-trained BEiT~\citep{bao2021beit}) by 0.3\% mIoU. Based on ViT-B, we fine-tune models for 160K iterations with AdamW following MAE and CAE. Tab.~\ref{tab:downstream_vit} shows that our approach consistently improves MIM methods (\textit{e.g.,} outperforms MAE and SimMIM by 0.9\% and 0.6\% mIoU).

\begin{table}[ht]
    \vspace{-1.5em}
    \setlength{\tabcolsep}{1.6mm}
    \centering
    \caption{Ablation of A$^2$MIM on IN-100 and IN-1K. $\rm{w/o} \ \omega$ denotes removing the re-weighting term $\omega$ in $\mathcal{L}_{freq}$ and $T (l^*)$ denotes adding the mask token $T$ to the optimal layer-$l^*$.}
    \label{tab:ablation_fft}
    \vspace{1pt}
\resizebox{\linewidth}{!}{
    \begin{tabular}{l|cccc}
    \toprule
    Backbones                                     & \multicolumn{2}{c}{ResNet-50} & ViT-S        & ViT-B       \\
    Datasets                                      & IN-100         & IN-1K        & IN-100       & IN-1K       \\
    Pre-training Epochs                           & 400 ep         & 100 ep       & 400 ep       & 400 ep      \\ \hline
    \gray{SimMIM}                                 & \gray{87.75}   & \gray{78.2}  & \gray{85.10} & \gray{83.1} \\
    $\mathcal{L}_{spa}$                           & 88.19          & 78.4         & 85.27        & 83.2        \\
    $+\mathcal{L}_{freq} \ \rm{w/o} \ \omega$     & 88.47          & 78.4         & 86.05        & 83.3        \\
    $+\mathcal{L}_{freq}$                         & 88.73          & 78.6         & 86.41        & 83.4        \\
\rowcolor{gray94} $+\mathcal{L}_{freq} + T (l^*)$ & \bf{88.86}     & \bf{78.8}    & \bf{86.62}   & \bf{83.5}   \\
    \bottomrule
    \end{tabular}
    }
    \vspace{-0.5em}
\end{table}

\vspace{-0.5em}
\subsection{Ablation Study}
\label{sec:exp_ablation}
We next verify the effectiveness of the proposed components. Ablation studies are conducted with ResNet-50 and ViTs on IN-100 and IN-1K using the fine-tuning protocol. Based on the modified baseline SimMIM ($\mathcal{L}_{spa}$), we first compare different mask token mechanisms: \textbf{Replacing} denotes the original way in most MIM methods, and \textbf{Addition} denotes our proposed way that adds the mask token to intermediate feature maps of the backbone. Replacing masked patches in input images by RGB mean value slightly improves the baseline SimMIM, especially for ResNet-50 (88.19 \textit{vs.} 87.75 on IN-100). Then, we verify the proposed $\mathcal{L}_{freq}$ in Tab.~\ref{tab:ablation_fft}. We find that simply using $\mathcal{L}_{freq}$ without the adaptive re-weighting $\omega$ (Eqn.~\ref{eq:reweight}) brings limited improvements as the frequency constraint to $\mathcal{L}_{spa}$, while employing $\omega$ further enhances performances by helping the model to learn more informative frequency components. Additionally, we visualize reconstruction results in Fig.~\ref{fig:ablation_rec_vit} to show the improvements brought by our proposed components. Refer to Appendix~\ref{app:ablation} and \ref{app:visualization} for more ablations and visualization results.

\begin{figure*}[t]  
    \vspace{-0.25em}
\centering
    \subfigtopskip=-0.5pt
    \subfigbottomskip=-0.5pt
    \subfigcapskip=-4pt
    \subfigure[]{\label{fig:in1k_mask_a}\includegraphics[height=0.211\linewidth,trim= 5 2 0 0,clip]{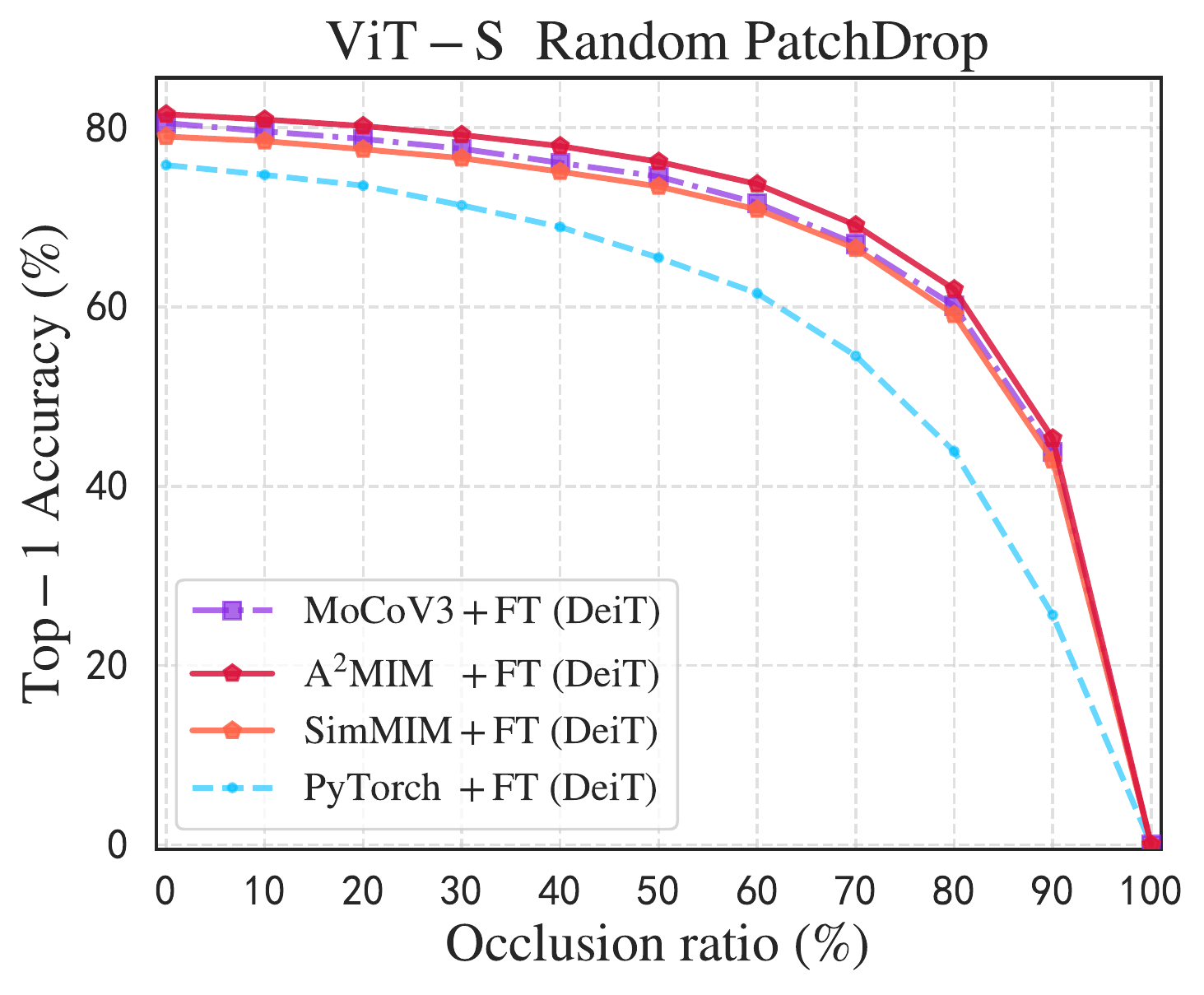}}
    \hspace{-0.25cm}
    \subfigure[]{\label{fig:in1k_mask_b}\includegraphics[height=0.211\linewidth,trim= 5 2 0 0,clip]{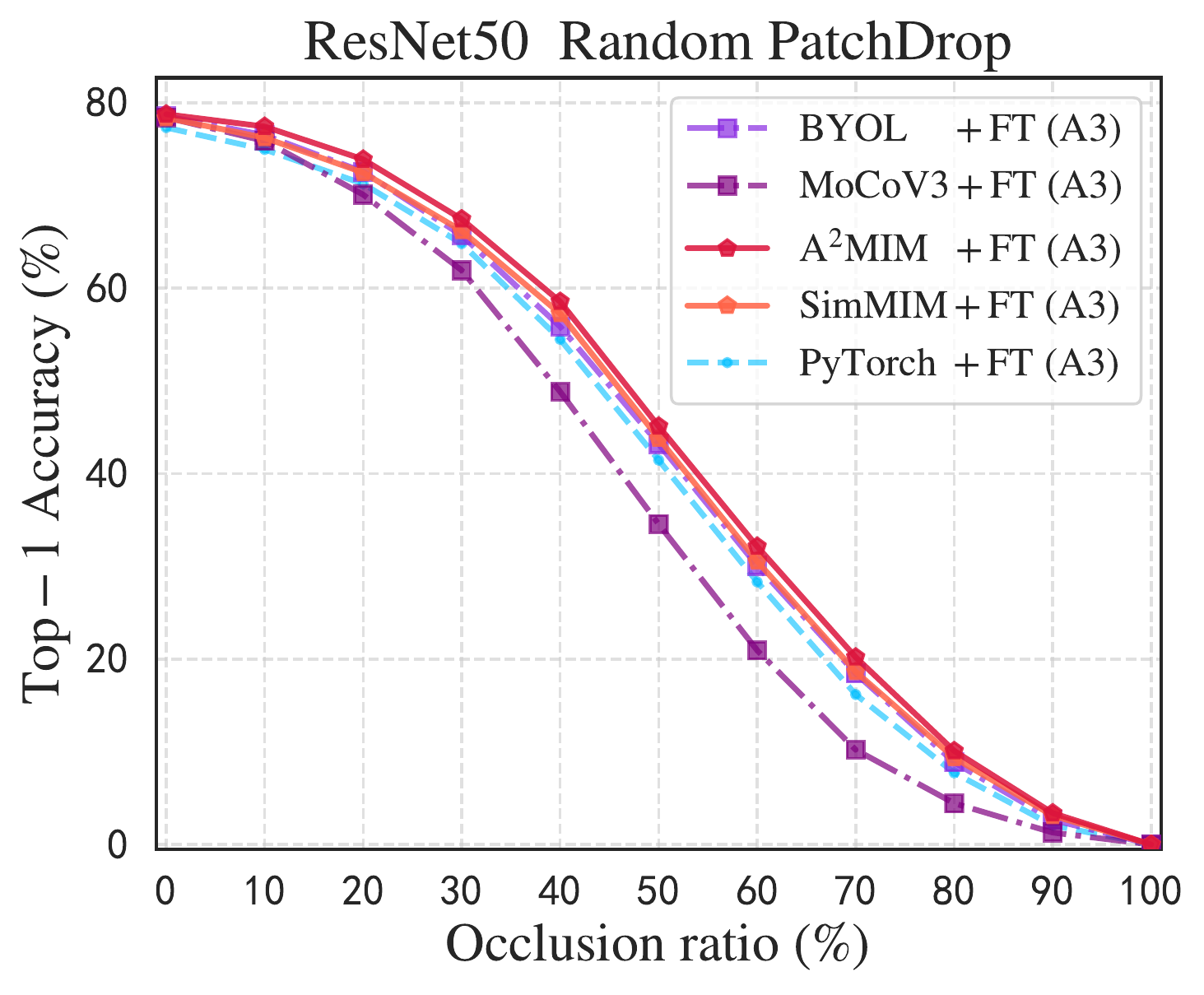}}
    \hspace{-0.25cm}
    \subfigure[]{\label{fig:in1k_mask_c}\includegraphics[height=0.211\linewidth,trim= 5 2 0 0,clip]{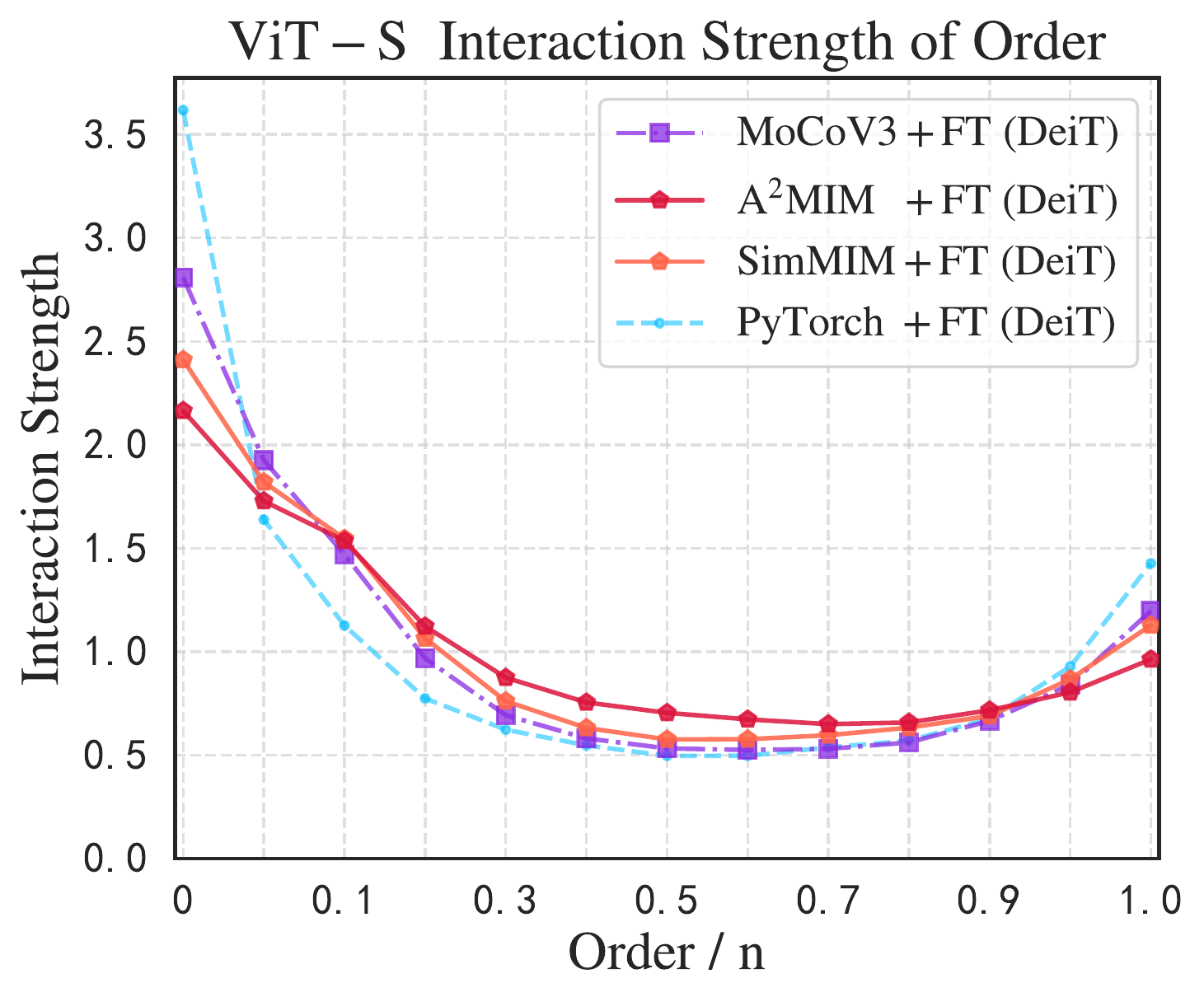}}
    \hspace{-0.25cm}
    \subfigure[]{\label{fig:in1k_mask_d}\includegraphics[height=0.211\linewidth,trim= 5 2 0 0,clip]{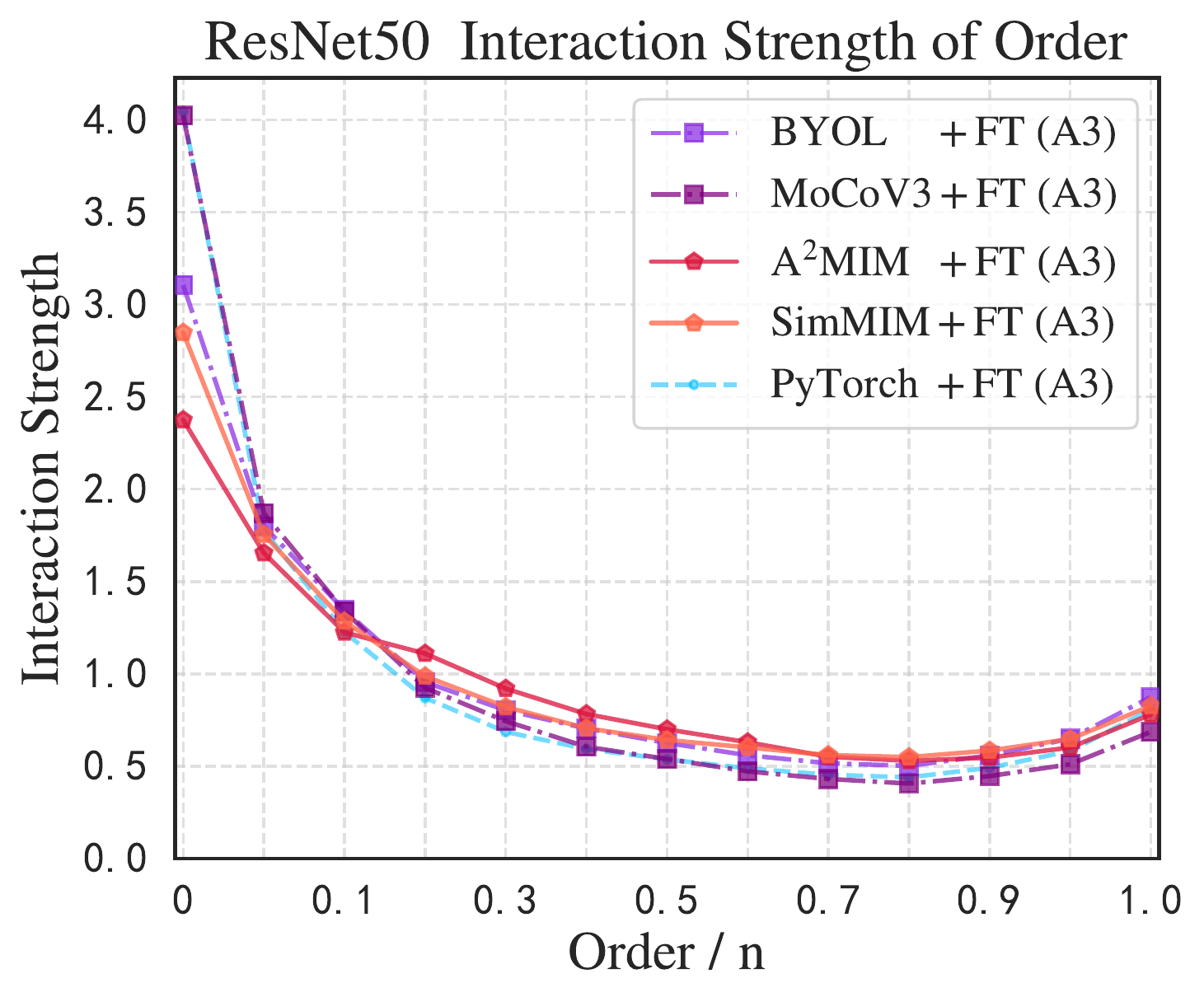}}
\vspace{-1.5em}
    \caption{
    Robustness and interaction of A$^2$MIM with ViT-S and ResNet-50 on ImageNet-1K. (a)(b): Robustness against different occlusion ratios of images is studied for A$^2$MIM and various methods. (c)(d): Distributions of the interaction strength $J^{(m)}$ are explored.}
    \label{fig:in1k_mask_interact}
    \vspace{-0.5em}
\end{figure*}

\begin{figure*}[t]  
\centering
    \subfigtopskip=-0.5pt
    \subfigbottomskip=-0.5pt
    \subfigcapskip=-4pt
    \subfigure[]{\label{fig:in1k_ep_a}\includegraphics[height=0.208\linewidth,trim= 0 10 0 5,clip]{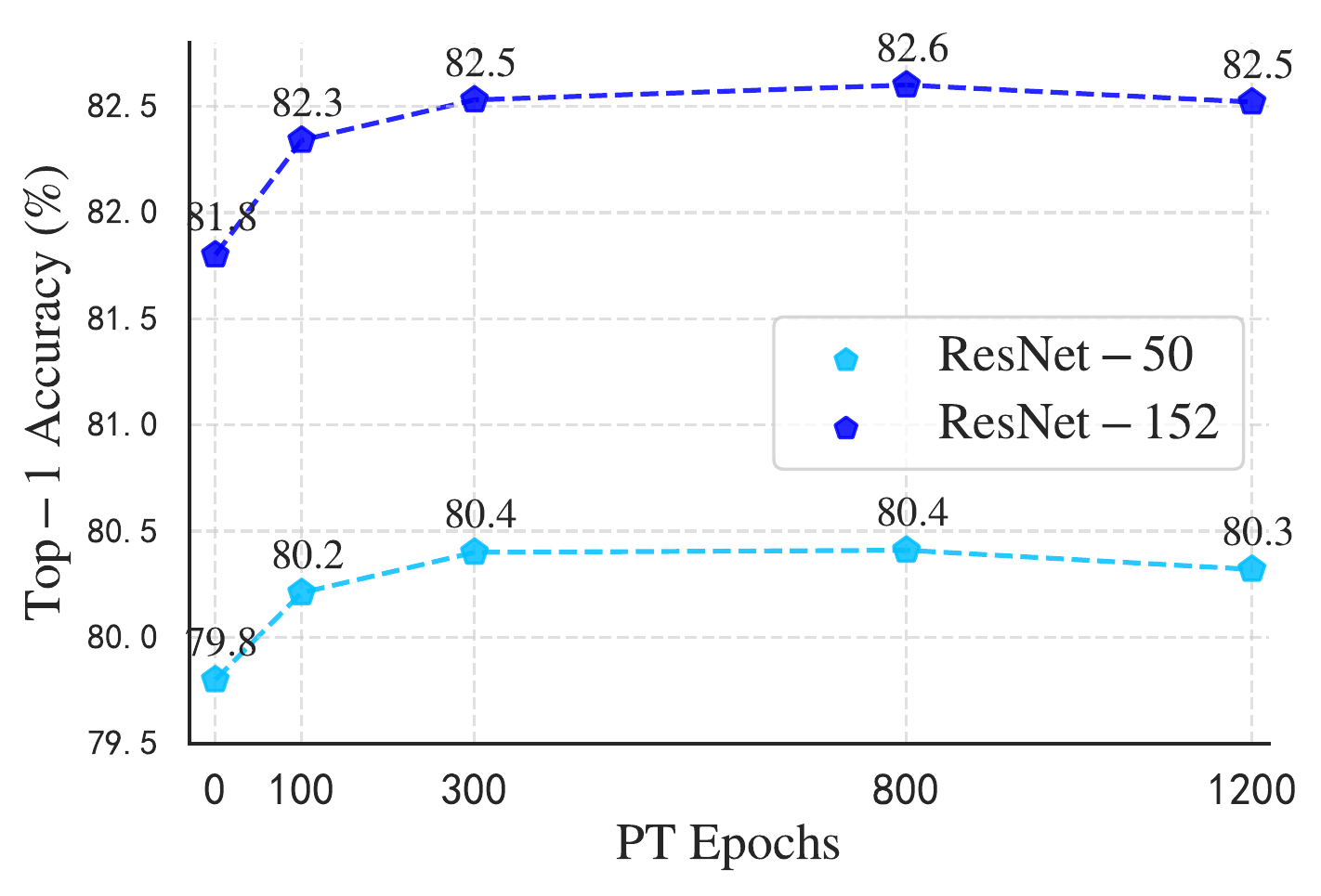}}
    \hspace{0.05cm}
    \subfigure[]{\label{fig:in1k_ep_b}\includegraphics[height=0.208\linewidth,trim= 0 10 0 5,clip]{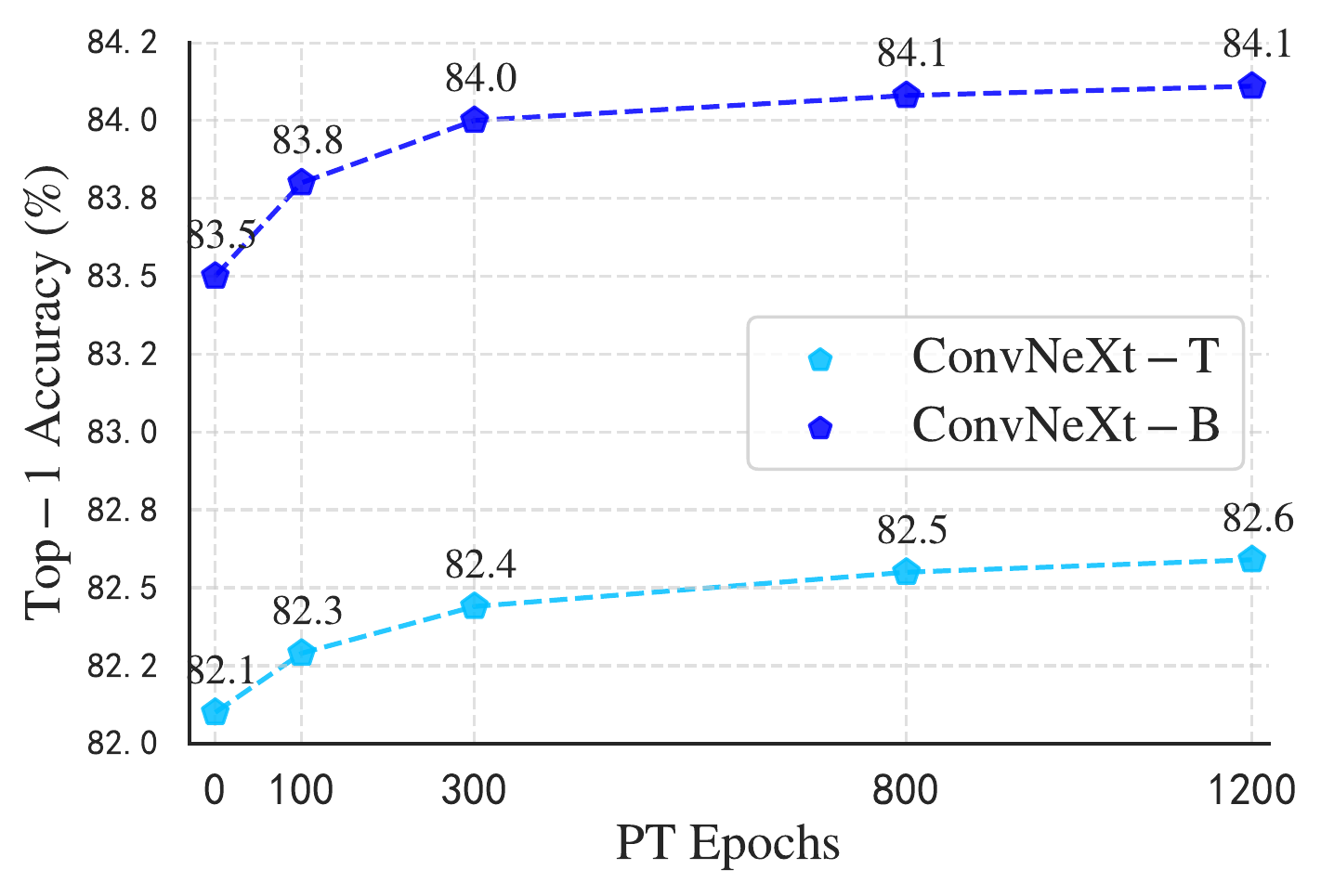}}
    \hspace{0.05cm}
    \subfigure[]{\label{fig:in1k_ep_c}\includegraphics[height=0.208\linewidth,trim= 0 10 0 5,clip]{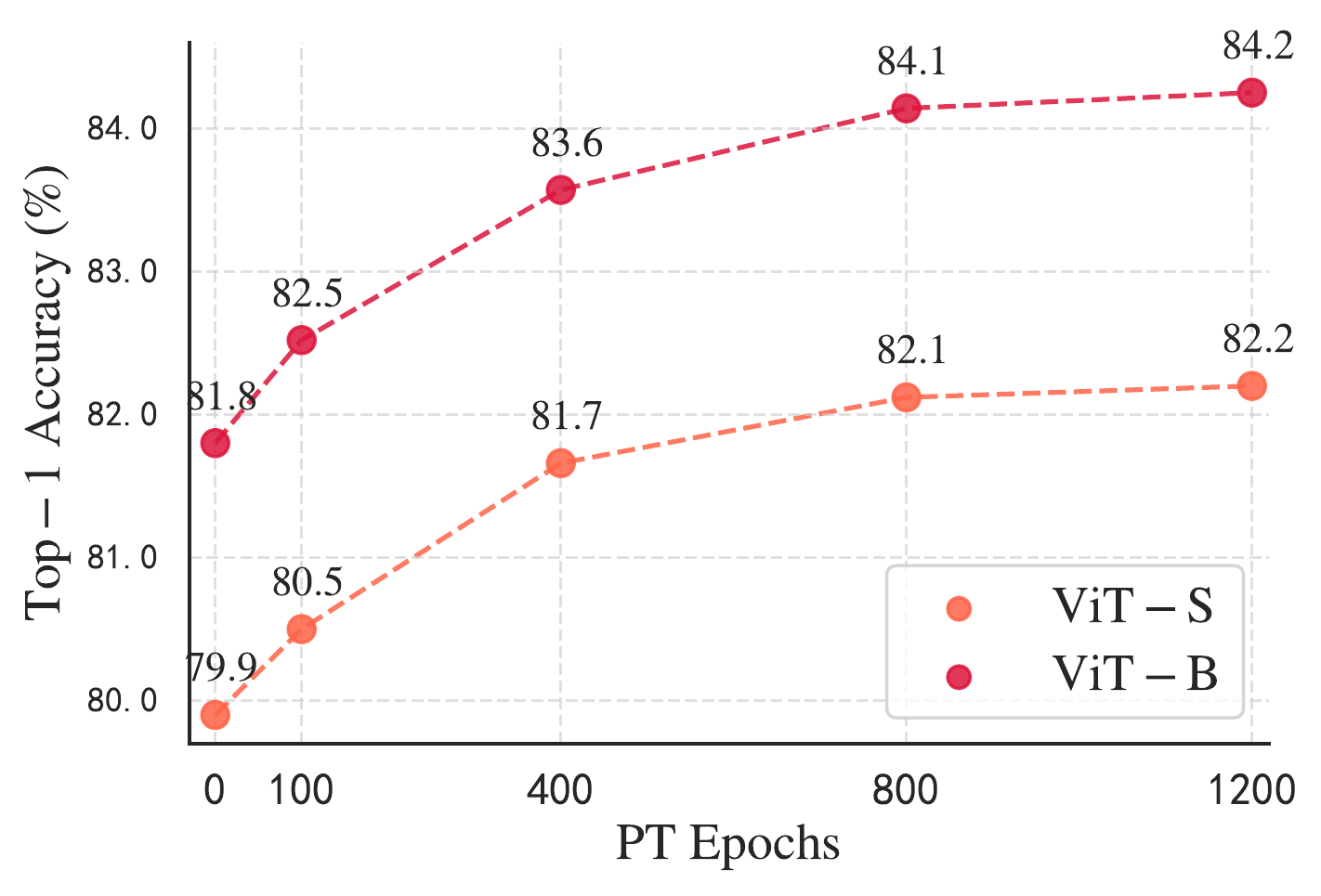}}
\vspace{-1.25em}
    \caption{Analysis of A$^2$MIM pre-training epochs and fine-tune performances with ResNet, ConvNeXt, and ViT models on ImageNet-1K. (a)(b) show CNN architectures obtain less performance gains and benefit less from longer pre-training from A$^2$MIM than ViTs in (c).}
    \label{fig:in1k_pt_ep}
    \vspace{-1.0em}
\end{figure*}

\subsection{Verification of A$^2$MIM Design Rules}
\label{sec:exp_abalysis}
We verify whether A$^2$MIM meets the intended design rules using the same experiment settings as Sec.~\ref{sec:exp_ablation} from two aspects. (i) \textbf{A$^2$MIM is generic to incorporate advanced components} proposed in previous works (\textit{e.g.}, complex decoders, advanced prediction targets). As for the decoder structure, we replace the original linear decoder with 2-layer MLP or Transformer decoders, but find limited improvements or degenerated performances (similar to SimMIM) in Tab.~\ref{tab:analysis_components}. Inspired by PVT.V2~\citep{Wang2022PVTv2}, we introduce a depth-wise (DW) convolution layer (\textbf{$w/$ DW}) to the MLP decoder (adding a $5\times 5$ DW layer in between) and the Transformer decoder (adding a $3\times 3$ DW layer in each FFN~\citep{Wang2022PVTv2}), which brings improvements compared to the linear decoder. As for the prediction target, we follow MaskFeat to change the RGB target to the HoG feature or the output feature from ViT-B/16 pre-trained by DINO~\citep{iccv2021dino}. Tab.~\ref{tab:analysis_components} shows that using advanced targets significantly improves the performance of $A^2$MIM for both ResNet-50 and ViT-B. Therefore, we can conclude \textit{A$^2$MIM is a generally applicable framework}.
(ii) \textbf{A$^2$MIM enhances occlusion robustness and middle-order interaction among patches} from experiments on IN-1K in Fig.~\ref{fig:in1k_mask_interact}. We analyze occlusion robustness and interaction strength of A$^2$MIM with ViT-S (pre-training 400-epoch) and ResNet-50 (pre-training 100-epoch) on ImageNet-1K, as shown in Fig.~\ref{fig:in1k_mask_interact}. Fig.~\ref{fig:in1k_mask_a} and \ref{fig:in1k_mask_b} shows that A$^2$MIM is more robust to occlusion than the baseline SimMIM and contrastive learning methods with both Transformers and CNNs. Meanwhile, we find that MIM methods learn more balanced interaction strength than both supervised and contrastive learning methods in Fig.~\ref{fig:in1k_mask_c} and \ref{fig:in1k_mask_d}. A$^2$MIM further improves SimMIM by capturing more middle-order interactions ($0.2n$ to $0.6n$) with Transformers and CNNs. Therefore, we can conclude that A$^2$MIM helps the model to learn better middle-order interactions between patches for more generalized visual representation.

\begin{table}[htb]
    \vspace{-1.5em}
    \setlength{\tabcolsep}{1.4mm}
    \centering
    \caption{Analysis of the scalability for advanced components.}
    \label{tab:analysis_components}
    \vspace{1pt}
\resizebox{\linewidth}{!}{
    \begin{tabular}{c|c|cc}
    \toprule
        & Module                                & ResNet-50                   & ViT-B                  \\ \hline
        & \gray{Linear}                         & \gray{78.8}                 & \gray{82.4}            \\
        & 2-layer MLP                           & 78.8                        & 82.4                   \\
Decoder & \cellcolor{gray94}2-layer MLP (w/ DW) & \cellcolor{gray94}\bf{78.9} & \cellcolor{gray94}82.5 \\
        & 2-layer Transformer                   & 78.6                        & 82.3                   \\
        & 2-layer Transformer (w/ DW)           & 78.8                        & \bf{82.6}              \\ \hline
        & \gray{RGB}                            & \gray{78.8}                 & \gray{82.4}            \\
Target  & HoG Feature                           & 78.9                        & 82.6                   \\
        & \cellcolor{gray94}DINO Feature        & \cellcolor{gray94}\bf{79.0} & \cellcolor{gray94}\bf{82.7} \\
    \bottomrule
    \end{tabular}
}
    \vspace{-1.0em}
\end{table}

\section{Conclusion and Limitation}
\label{sec:conclusion}
In this paper, we attempted to answer the question of what is learned during MIM pre-training. We adopted multi-order interactions to study the interaction order among image patches. We discovered that MIM essentially teaches the network to learn middle-order interactions among image patches for more complex feature extraction regardless of the network architecture. Based on our findings, we further proposed a general MIM framework A$^2$MIM that is compatible with both Transformers and CNNs. Besides a proposed novel masking mechanism, we also proposed a loss in the Fourier domain to enhance the middle-order interaction among patches. Experimental results showed that our proposed framework improves the representations learned for CNNs and Transformers, yielding superior performance than prior methods on various downstream tasks.

Meanwhile, we list two limitations of A$^2$MIM, as shown in Figure~\ref{fig:in1k_pt_ep}. (i) CNNs architectures benefit less from A$^2$MIM pre-training compared to ViTs, \textit{e.g.}, ResNet and ConvNeXt gain around 1\% Acc while ViTs obtain more than 2\% gains. We hypothesize that the inductive bias of CNNs limits the learning of middle-order interactions induced by MIM. (ii) ViTs benefit more with longer pre-training, while no significant gain is observed for CNNs after 300 epochs pre-training. Figure~\ref{fig:in1k_ep_a} shows that ResNet-50/152 obtains limited or negative performance gains for pre-training of 800 epochs or more. We hope our work could inspire the community to further promote self-supervised pre-training.

\section*{Acknowledgement}
This work was supported by National Key R\&D Program of China (No. 2022ZD0115100), National Natural Science Foundation of China Project (No. U21A20427), and Project (No. WU2022A009) from the Center of Synthetic Biology and Integrated Bioengineering of Westlake University.





\bibliography{reference}
\bibliographystyle{icml2023}


\clearpage
\renewcommand\thefigure{A\arabic{figure}}
\renewcommand\thetable{A\arabic{table}}
\setcounter{table}{0}
\setcounter{figure}{0}

\appendix

\section{Details of Comparison Experiments}
\label{app:comparison_exp}
This section provides experimental details for Sec.~\ref{sec:exp}, \textit{e.g.,} pre-training and evaluation on ImageNet-1K and transferring learning settings on downstream tasks. Experiment results and models are available at \url{https://github.com/Westlake-AI/A2MIM}.

\vspace{-0.25em}
\subsection{ImageNet-1K Experiments}
\paragraph{Pre-training.}
The default settings of A$^2$MIM for CNNs and ViTs are provided in Tab.~\ref{tab:app_pt_config}, following SimMIM~\citep{xie2021simmim}. We use AdamW~\citep{iclr2019AdamW} optimizer with the cosine scheduler and the linear learning rate scaling rule~\citep{2017msgd}: \textit{lr} = \textit{base\_lr}$\times$batchsize / 2048. Similar to current MIM methods, we only employ \textit{RandomResizedCrop} with the scale of $(0.67,1.0)$ or $(0.8,1.0)$ and \textit{RandomFlip}, while do not require other complex augmentations (\textit{e.g.,} Rand Augment~\citep{cubuk2020randaugment}, mixups~\citep{zhang2017mixup,yun2019cutmix, eccv2022AutoMix, li2021samix}, or stochastic depth~\citep{eccv2016droppath}) during pre-training. As for ResNet and ConvNeXt models, we adopt Cosine learning rate decay for 100/300 and 800 epochs pre-training. As for ViTs, we use a similar Cosine decay when pre-training epochs less than 400 while using Step decay (the learning rate multiplied by $0.1$ at 700-epoch) for 800-epoch pre-training.

\vspace{-1.0em}
\paragraph{End-to-end fine-tuning.}
As shown in Tab.~\ref{tab:app_ft_config}, our fine-tuning settings follow common practices of supervised image classification on ImageNet-1K. For ViT architectures, the pre-trained model is fine-tuned for 200 epochs using the BEiT~\citep{bao2021beit} version of DeiT~\citep{touvron2021training} training recipe to fully explore the performance, which employs AdamW~\citep{iclr2019AdamW} optimizer with the cross-entropy (CE) loss and layer-wise learning rate decay.
For CNNs, we adopt RSB A3~\citep{wightman2021rsb} setting for 100-epoch fine-tuning, which employs LAMB~\citep{iclr2020lamb} optimizer with the binary cross-entropy (BCE) loss and smaller training resolutions. To fully explore the PT performances of CNNs, we also apply 300-epoch fine-tuning with RSB A2~\citep{wightman2021rsb} and ConvNeXt~\citep{cvpr2022convnext} training settings for ResNet and ConvNeXt models. Notice that the default drop depth rates of ResNet-50/101/152/200 and ConvNeXt-T/S/B are 0.05/0.1/0.15/0.2 and 0.1/0.3/0.4 in 300-epoch fine-tuning. The learning rates and drop depth can also be tuned for different PT methods.

\vspace{-0.25em}
\subsection{Object Detection and Segmentation on COCO}
We adopt Mask-RCNN~\citep{2017iccvmaskrcnn} to perform transfer learning to object detection and semantic segmentation on COCO~\citep{eccv2014MSCOCO} using MMDetection{\footnote{\url{https://github.com/open-mmlab/mmdetection}}} and Detectron2{\footnote{\url{https://github.com/facebookresearch/detectron2}}} code bases.
For evaluation on ResNet-50, we follow MoCo~\citep{cvpr2020moco} and fine-tune Mask R-CNN with the pre-trained ResNet-50-C4 backbone with SGD optimizer using 2$\times$ schedule (24 epochs). For evaluation of ViTs, we follow MAE~\citep{he2021masked} and CAE~\citep{2022cae}, which apply the pre-trained ViT backbone and an FPN neck~\citep{cvpr2017fpn} in Mask R-CNN. The model is fine-tuned by AdamW optimizer with 1$\times$ schedule (12 epochs). For a fair comparison, we follow~\citep{bao2021beit,xie2021simmim} to turn on relative position bias in ViT~\citep{dosovitskiy2020image} during both pre-training and transfer learning, initialized as zero, and the learning rate can be tuned for different PT methods.

\begin{table}[ht]
    \vspace{-1.0em}
    \setlength{\tabcolsep}{0.9mm}
    \centering
    \caption{ImageNet-1K pre-training settings of A$^2$MIM for ResNet/ConvNeXt and ViT/Swin models.}
    \label{tab:app_pt_config}
\resizebox{\linewidth}{!}{
    \begin{tabular}{l|cc}
    \toprule
    Configuration           & ResNet / ConvNeXt               & ViT / Swin                      \\ \hline
    Pre-training resolution & $224\times 224$                 & $224\times 224$                 \\
    Mask patch size         & $32\times 32$                   & $32\times 32$                   \\
    Mask ratio              & 60\%                            & 60\%                            \\
    Optimizer               & AdamW                           & AdamW                           \\
    Base learning rate      & $1.2\times 10^{-3}$             & $4\times 10^{-4}$               \\
    Weight decay            & 0.05                            & 0.05                            \\
    Optimizer momentum      & $\beta_1, \beta_2{=}0.9, 0.999$ & $\beta_1, \beta_2{=}0.9, 0.999$ \\
    Batch size              & 2048                            & 2048                            \\
    Learning rate schedule  & Cosine                          & Step / Cosine                   \\
    Warmup epochs           & 10                              & 10                              \\
    RandomResizedCrop       & [0.8, 1]                        & [0.67, 1]                       \\
    Rand Augment            & \xmarkg                         & \xmarkg                         \\
    Stochastic Depth        & \xmarkg                         & \xmarkg                         \\
    Gradient Clipping       & \xmarkg                         & max norm$=5$                    \\
    PT epochs               & 100 / 300 / 800                 & 300 / 800                       \\
    \bottomrule
    \end{tabular}
    }
    \vspace{-1.0em}
\end{table}

\begin{table}[ht]
    \vspace{-0.5em}
    \setlength{\tabcolsep}{0.7mm}
    \centering
    \caption{ImageNet-1K fine-tuning recipes of ViT, RSB A2/A3, and ConvNeXt architectures. Here we take ViT-B, ResNet-50, and ConvNeXt-T as examples.}
    \label{tab:app_ft_config}
\resizebox{\linewidth}{!}{
\begin{tabular}{l|cccc}
	\toprule
    Configuration              & ViT                 & RSB A2              & RSB A3            & ConvNeXt          \\ \hline
    FT epochs                  & 200                 & 300                 & 100               & 300               \\
    Training resolution        & 224                 & 224                 & 160               & 224               \\
    Testing resolution         & 224                 & 224                 & 224               & 224               \\
    Testing crop ratio         & 0.875               & 0.95                & 0.95              & 0.875             \\
    Optimizer                  & AdamW               & LAMB                & LAMB              & AdamW             \\
    Base learning rate         & $1\times 10^{-2}$   & $5\times 10^{-3}$   & $8\times 10^{-3}$ & $4\times 10^{-3}$ \\
    Layer-wise decay           & 0.65                & \xmarkg             & \xmarkg           & \xmarkg           \\
    Weight decay               & 0.05                & 0.02                & 0.02              & 0.05              \\
    Batch size                 & 1024                & 2048                & 2048              & 4096              \\
    Learning rate schedule     & Cosine              & Cosine              & Cosine            & Cosine            \\
    Warmup epochs              & 20                  & 5                   & 5                 & 20                \\
    Label smoothing $\epsilon$ & 0.1                 & \xmarkg             & \xmarkg           & 0.1               \\
    Stochastic depth           & 0.1                 & 0.05                & \xmarkg           & 0.1               \\
    Gradient clipping          & 5.0                 & \xmarkg             & \xmarkg           & \xmarkg           \\
    Rand Augment               & (9, 0.5)            & (7, 0.5)            & (6, 0.5)          & (9, 0.5)          \\
    Mixup alpha                & 0.8                 & 0.1                 & 0.1               & 0.8               \\
    CutMix alpha               & 1.0                 & 1.0                 & 1.0               & 1.0               \\
    EMA decay                  & 0.99996             & \xmarkg             & \xmarkg           & 0.9999            \\
    Loss function              & CE loss             & BCE loss            & BCE loss          & CE loss           \\
    \bottomrule
    \end{tabular}
    }
    \vspace{-0.5em}
\end{table}

\begin{figure*}[t]  
    \vspace{-0.5em}
\centering
    \subfigtopskip=-0.5pt
    \subfigbottomskip=-0.5pt
    \subfigcapskip=-4pt
    \subfigure[]{\label{fig:in100_mask_b}\includegraphics[height=0.211\linewidth,trim= 5 0 0 0,clip]{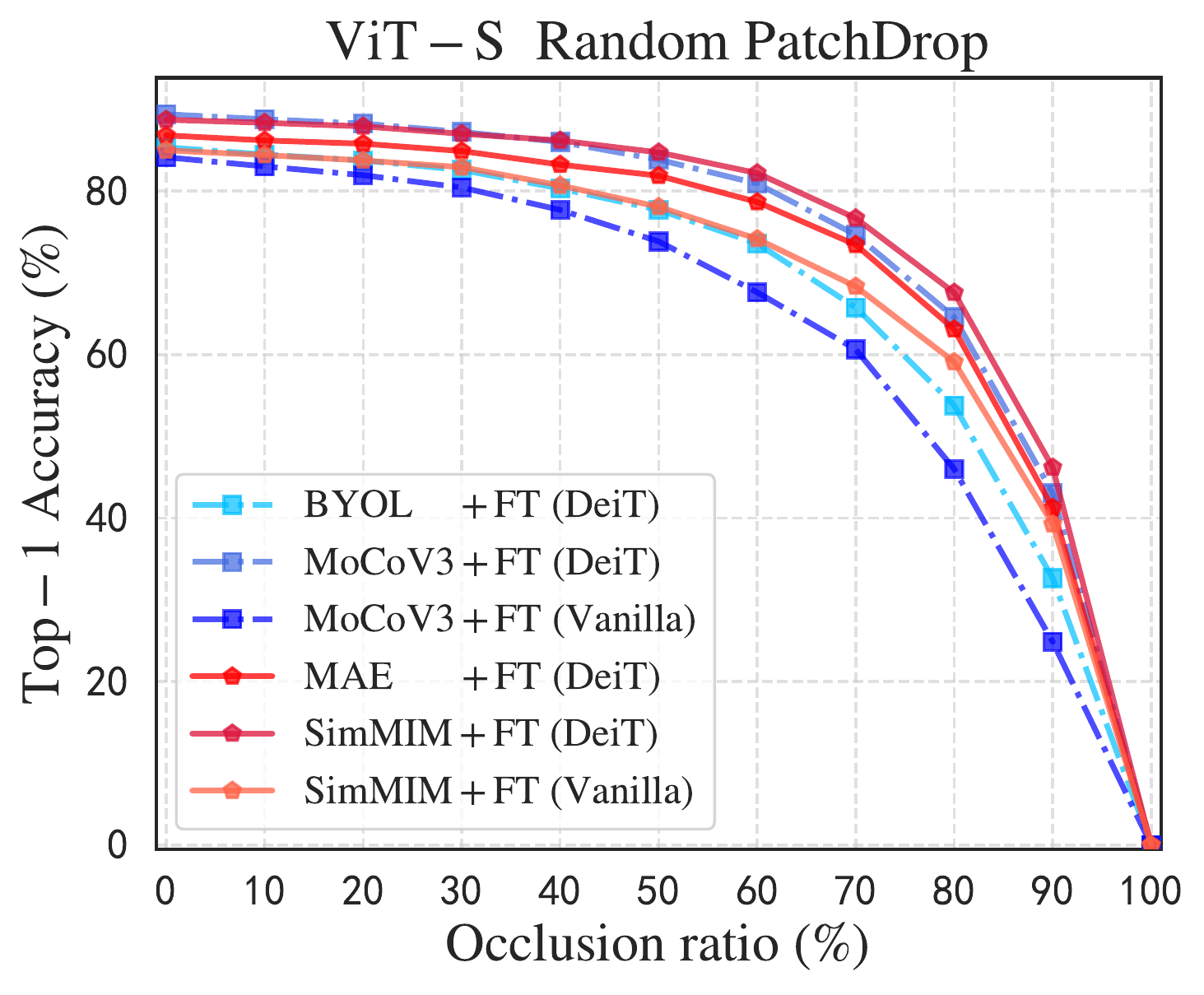}}
    \hspace{-0.25cm}
    \subfigure[]{\label{fig:in100_mask_d}\includegraphics[height=0.211\linewidth,trim= 5 0 0 0,clip]{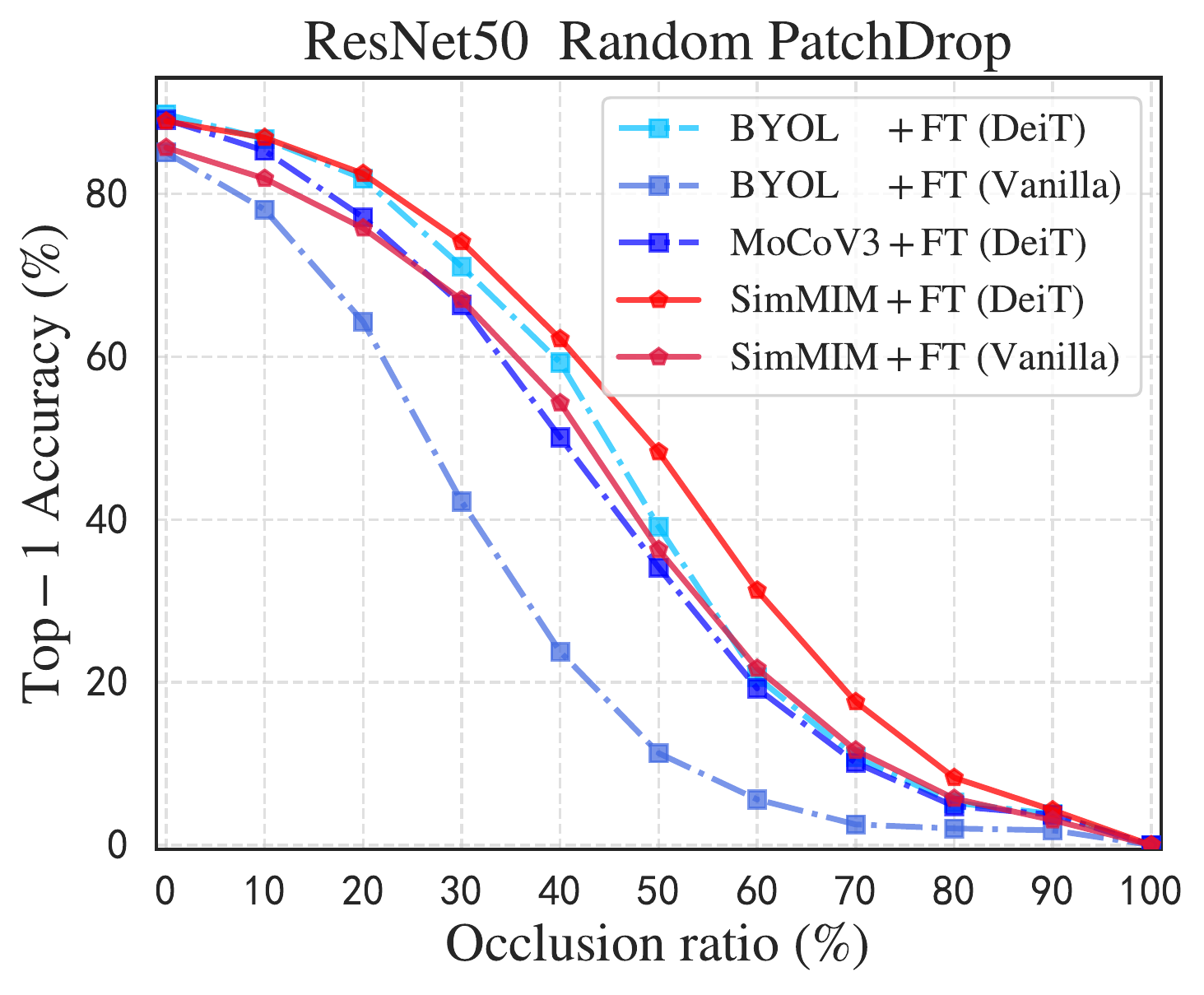}}
    \hspace{-0.25cm}
    \subfigure[]{\label{fig:in100_mask_f}\includegraphics[height=0.211\linewidth,trim= 5 0 0 0,clip]{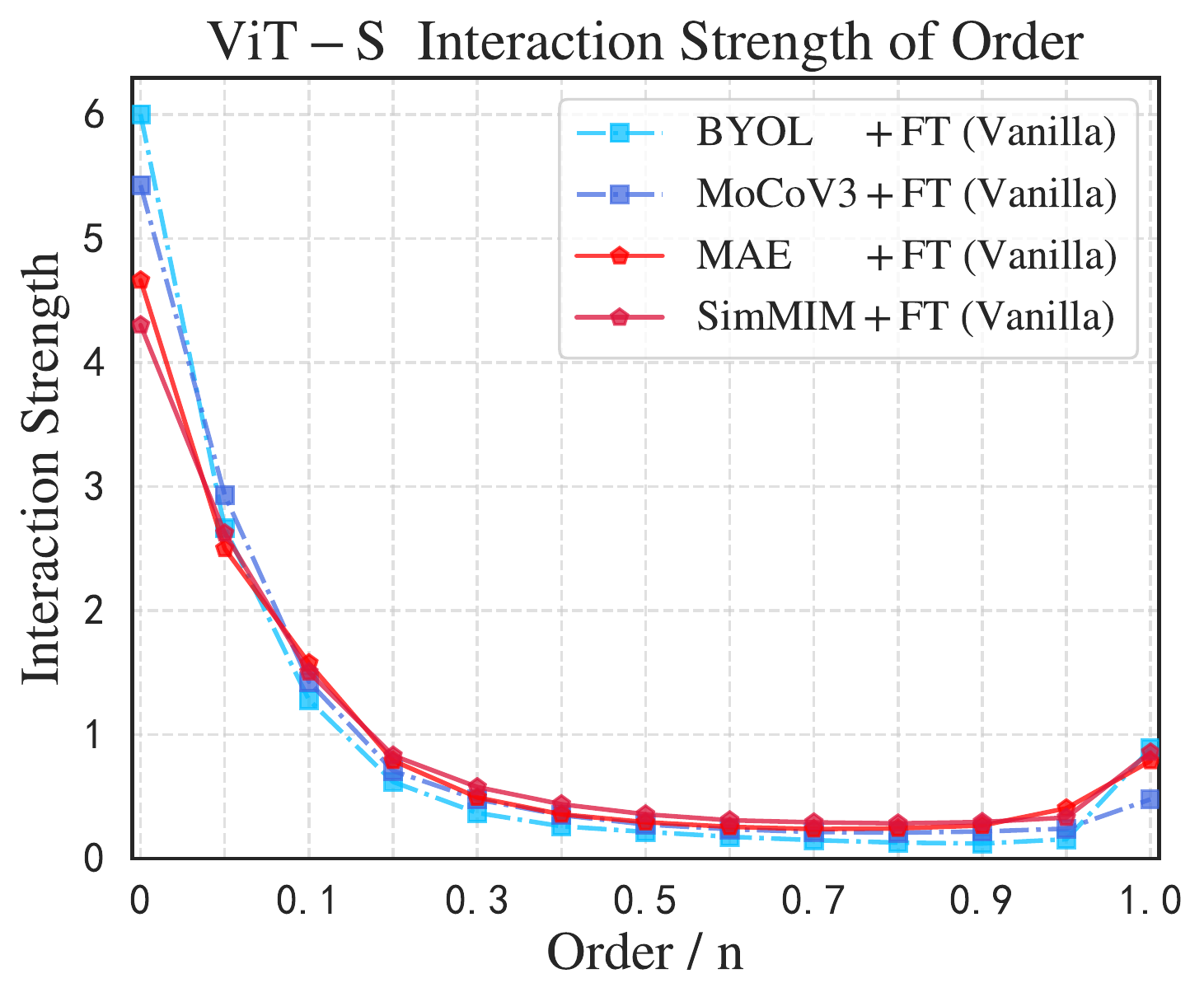}}
    \hspace{-0.25cm}
    \subfigure[]{\label{fig:in100_mask_h}\includegraphics[height=0.211\linewidth,trim= 5 0 0 0,clip]{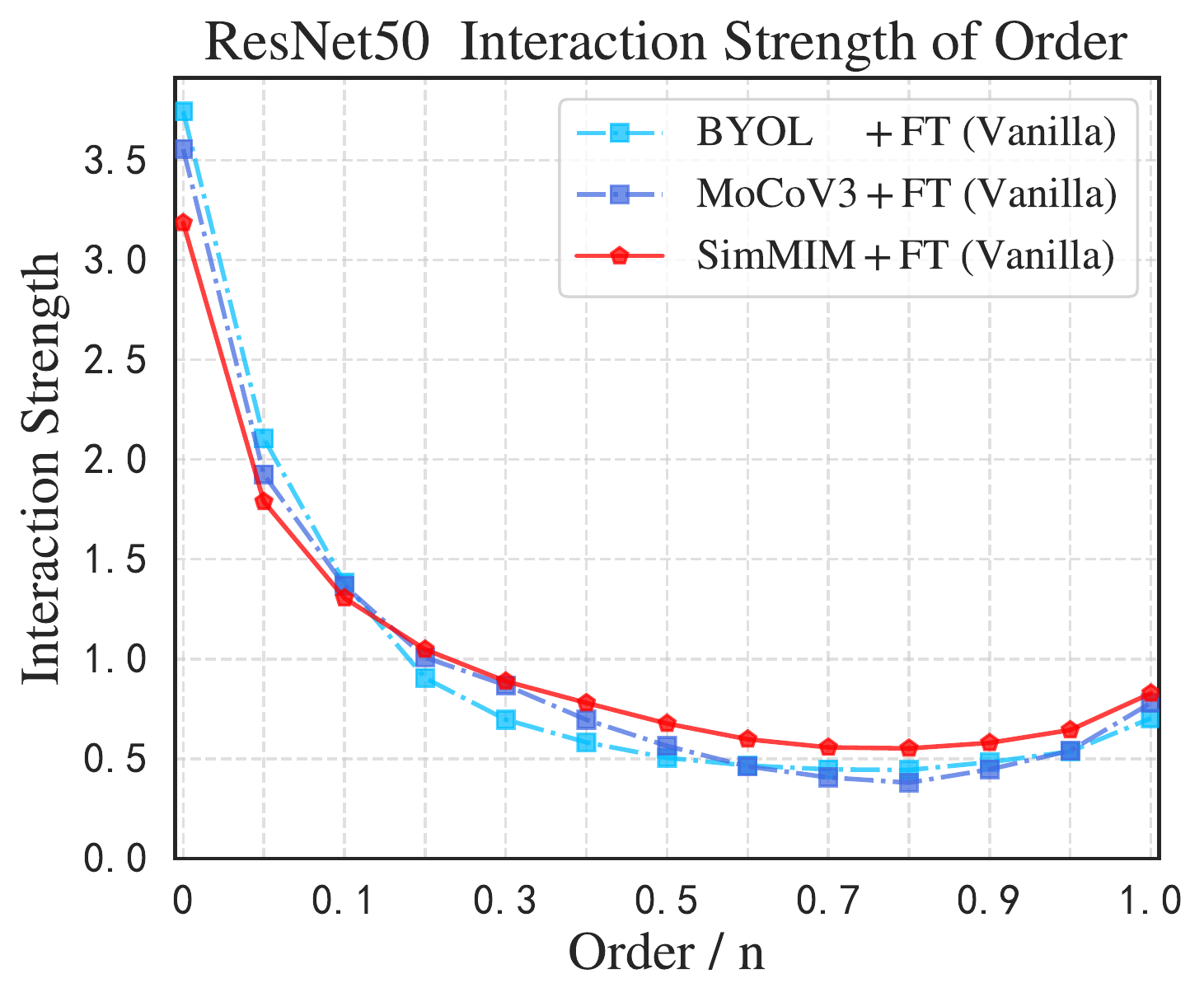}}
\vspace{-0.5em}
    \caption{(a)(b): Occlusion robustness against different occlusion ratios of images (CL \textit{vs.} MIM) is studied for both ViT-S and ResNet-50 on ImageNet-100. (c)(d): Distributions of the interaction strength $J^{(m)}$ (CL \textit{vs.} MIM) are explored for both ViT-S and ResNet-50 on ImageNet-100. The label indicates the pre-training method $+$ fine-tuning augmentation used, random stands for random weight initialization.}
    \label{fig:in100_mask_cl_vs_mim}
    \vspace{-0.5em}
\end{figure*}

\begin{figure*}[ht]  
\centering
    \subfigtopskip=-0.5pt
    \subfigbottomskip=-0.5pt
    \subfigcapskip=-4pt
    \subfigure[]{\label{fig:in100_mask_salient_a}\includegraphics[height=0.211\linewidth,trim= 5 0 0 0,clip]{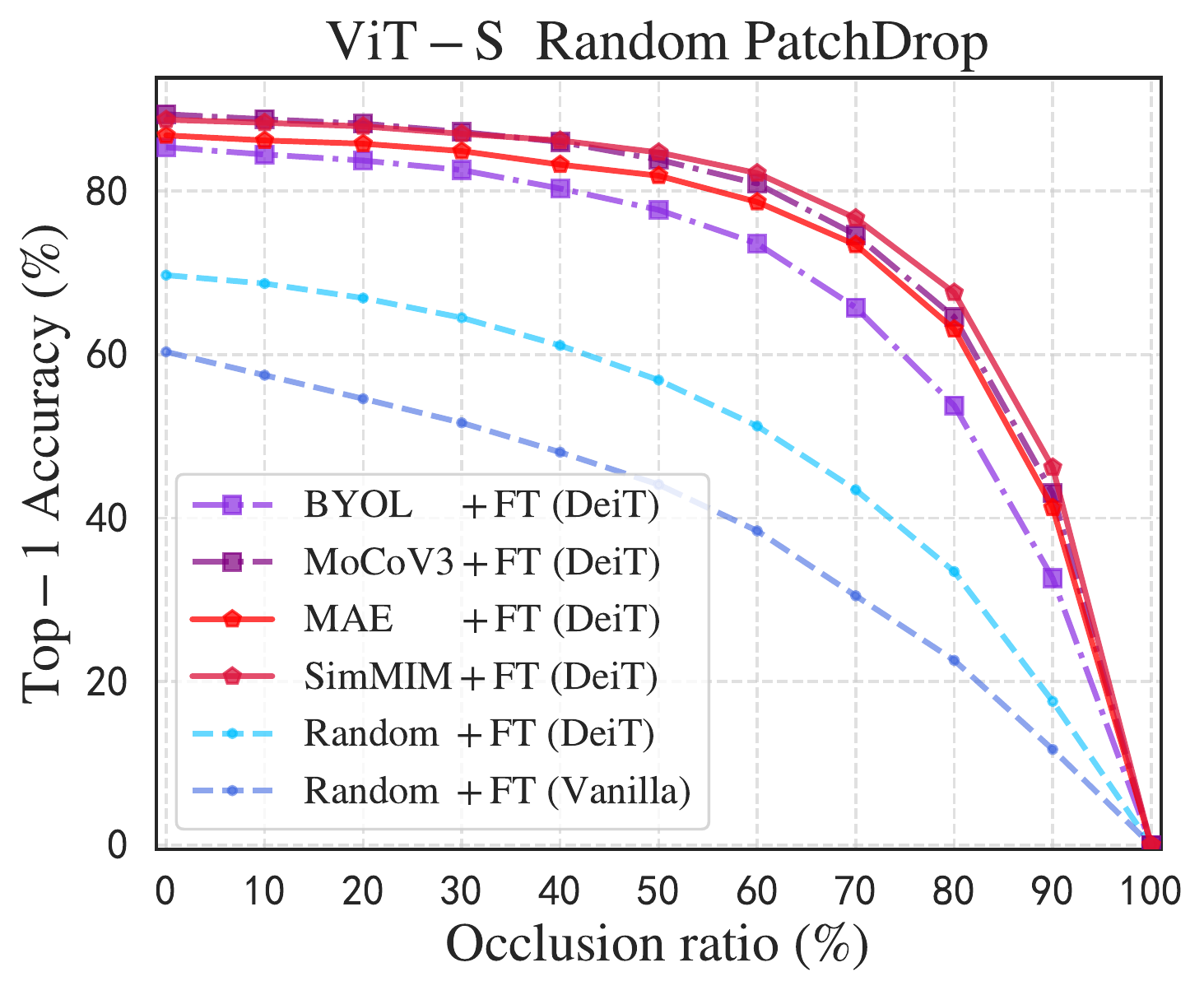}}
    \hspace{-0.25cm}
    \subfigure[]{\label{fig:in100_mask_salient_b}\includegraphics[height=0.211\linewidth,trim= 5 0 0 0,clip]{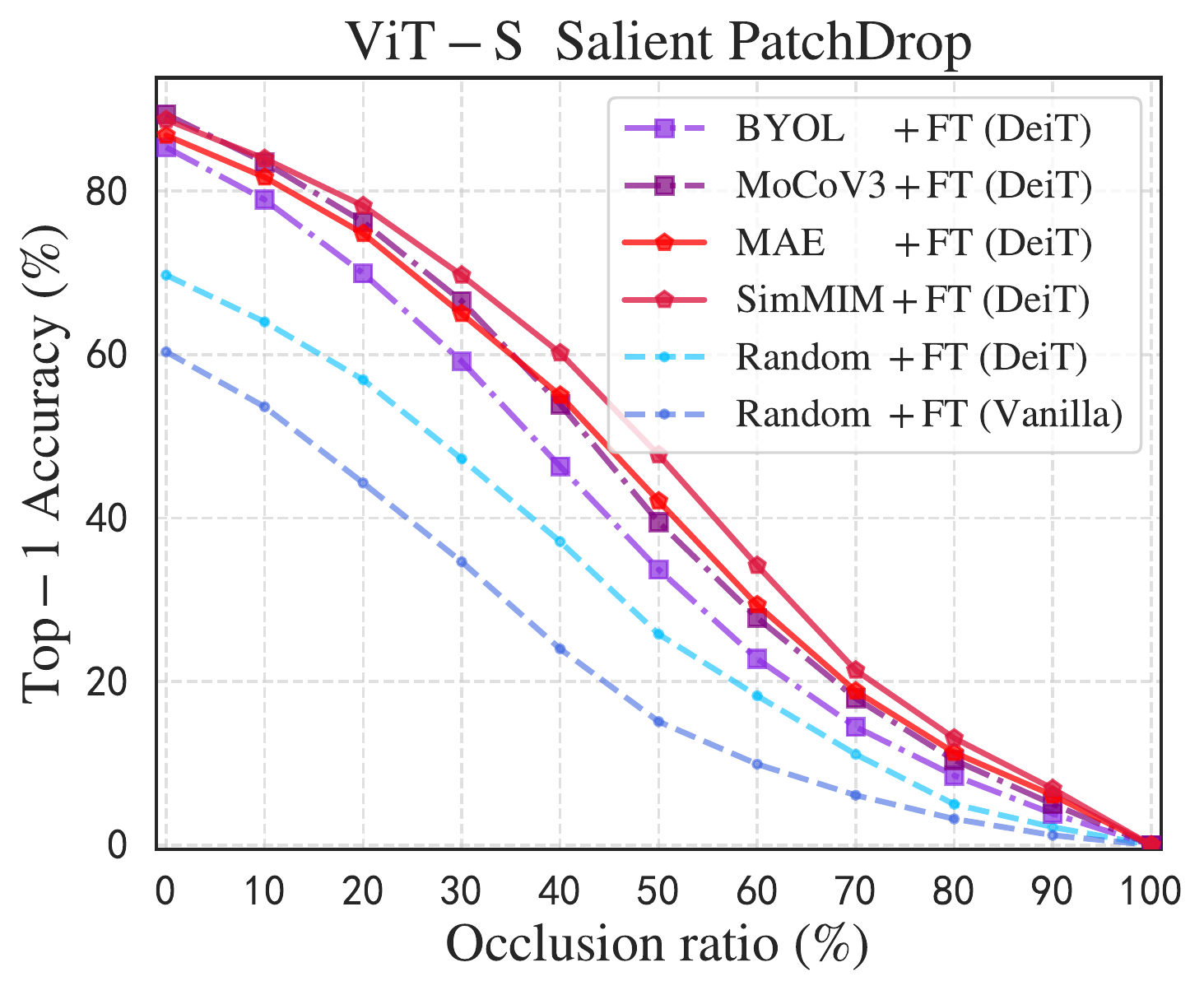}}
    \hspace{-0.25cm}
    \subfigure[]{\label{fig:in100_mask_salient_c}\includegraphics[height=0.211\linewidth,trim= 5 0 0 0,clip]{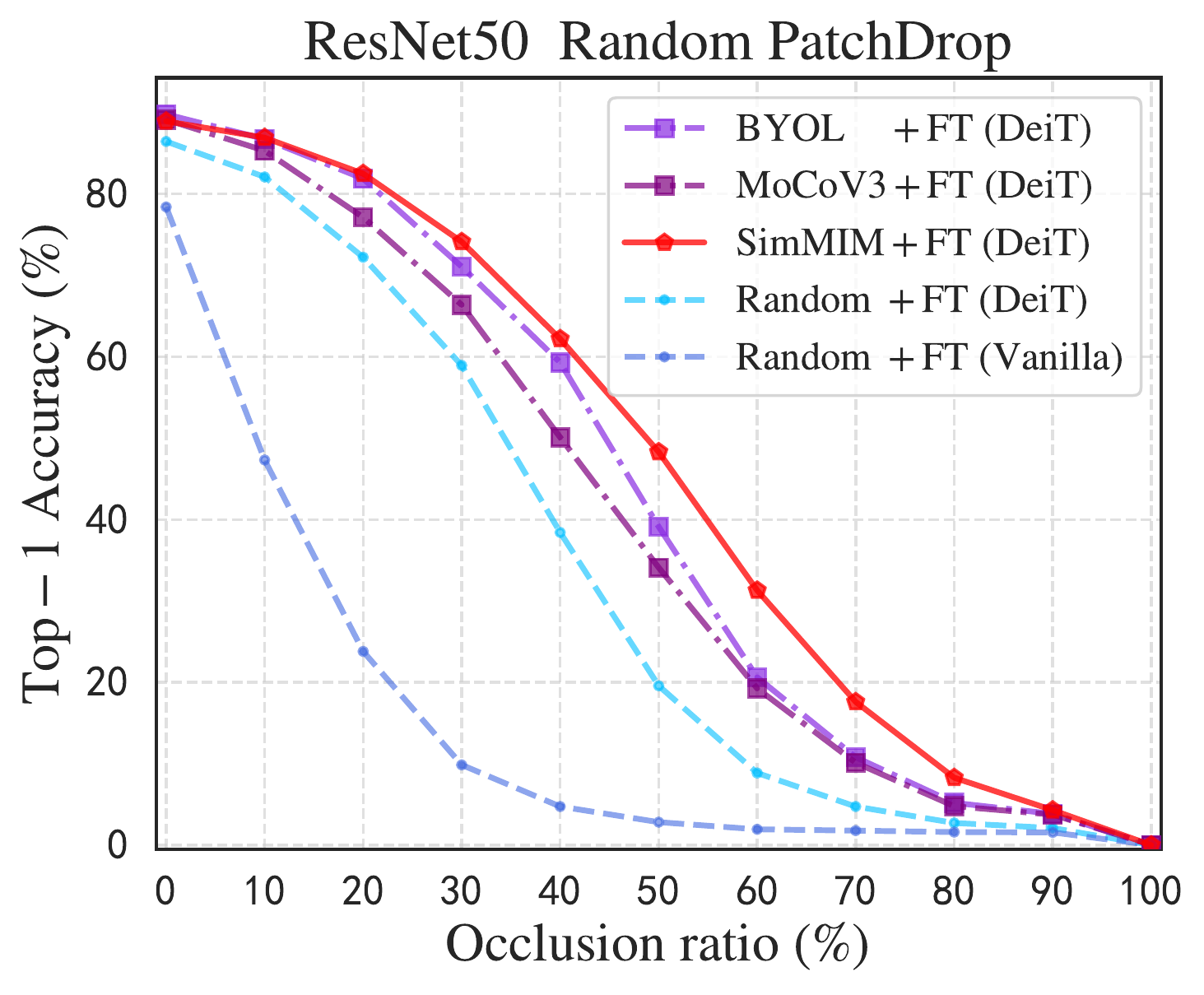}}
    \hspace{-0.25cm}
    \subfigure[]{\label{fig:in100_mask_salient_d}\includegraphics[height=0.211\linewidth,trim= 5 0 0 0,clip]{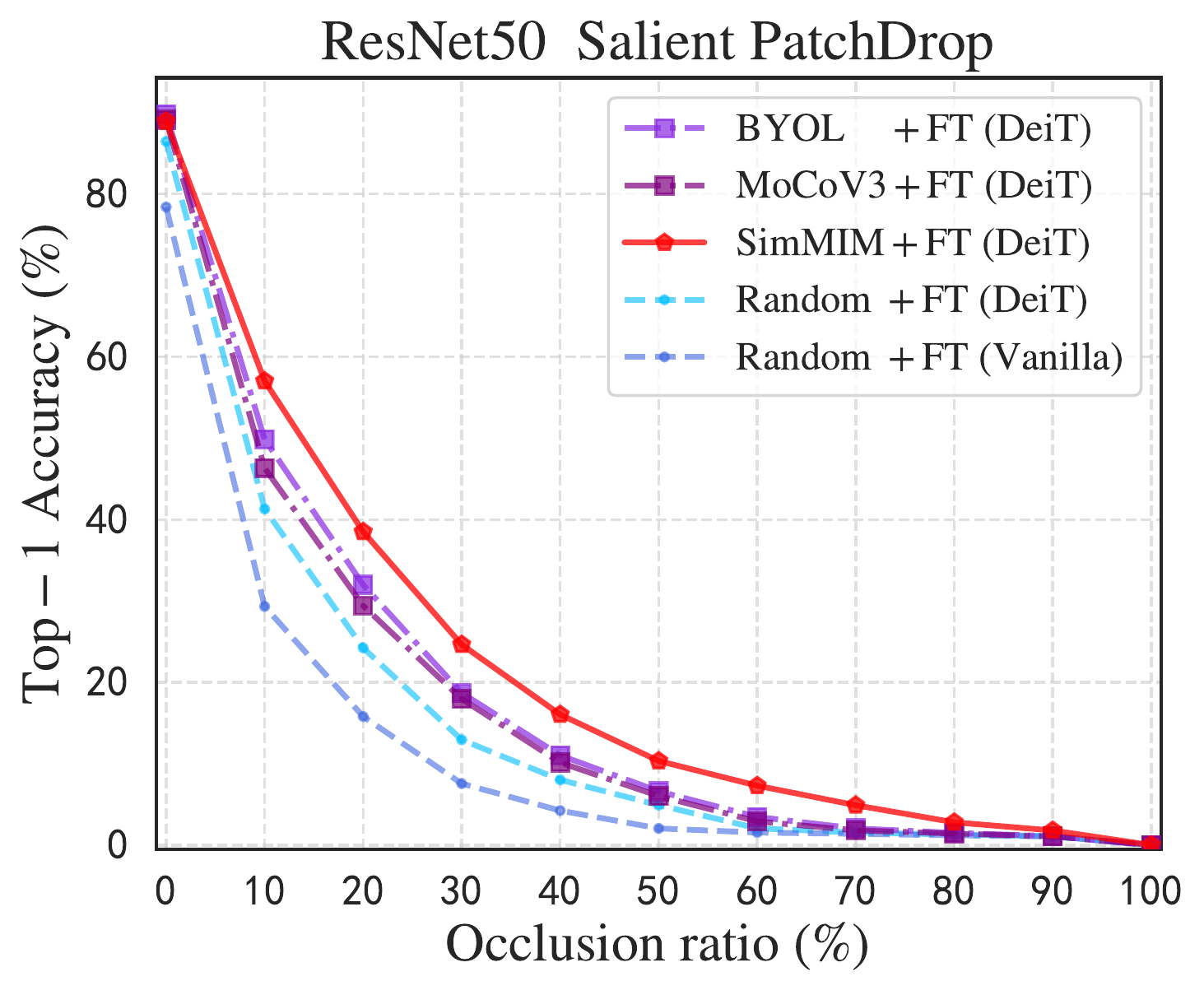}}
\vspace{-0.75em}
    \caption{
    Occlusion robustness against various random or salient occlusion ratios of images is studied in (a)(b) for ViT-S, and (c)(d) for ResNet-50 using various experimental settings on ImageNet-100. The label indicates the pre-training method $+$ fine-tuning setting used, random stands for random weight initialization.}
    \label{fig:app_in100_mask_salient}
    \vspace{-0.5em}
\end{figure*}

\subsection{Semantic Segmentation on ADE-20K}
We adopt UperNet~\citep{eccv2018upernet} to perform transfer learning to semantic segmentation on ADE-20K and use the semantic segmentation implementation in MMSegmentation{\footnote{\url{https://github.com/open-mmlab/mmsegmentation}}}. We initialize the FCN~\citep{tpami2017fcn} or UperNet~\citep{eccv2018upernet} using the pre-trained backbones (ResNet-50 or ViTs) on ImageNet-1K. For ViTs, we fine-tune end-to-end for 160K iterations with AdamW and a batch size of 16. We search a optimal layer-wise decay from \{0.8, 0.9\} and search optimal a learning rate from \{$1\times 10^{-4}, 2\times 10^{-4}, 3\times 10^{-4}$\} for all competitors. Similar to fine-tuning settings on COCO, we use relative position bias in ViT~\citep{dosovitskiy2020image} during both pre-training and transfer learning as~\citep{bao2021beit,xie2021simmim}. For ResNet-50, we follow MoCo~\citep{cvpr2020moco}, \textit{i.e.,} all CNN models are fine-tuned for 160K iterations by SGD optimizer with the momentum of 0.9 and a batch size of 16.

\section{Empirical Experiments}
\label{app:empirical_exp}
This section provides background information and experimental details for Sec.~\ref{sec:empirical}, and additional results of occlusion robustness evaluation and multi-order interaction strength.

\subsection{Occlusion Robustness}
\label{app:robustness}
In Sec.~\ref{sec:occlusion_robustness}, we analyze robustness against occlusion for models pre-trained and fine-tuned on ImageNet-100 (a subset on ImageNet-1K divided by~\citep{eccv2020CMC}) using the official implementation{\footnote{\url{https://github.com/Muzammal-Naseer/Intriguing-Properties-of-Vision-Transformers}}} provided by \citet{naseer2021intriguing}. Both MIM and contrastive-based methods are pre-trained 400 epochs on ImageNet-100 using their pre-training settings on ImageNet-1K. We adopt the fine-tuning training recipe as DeiT in Tab.~\ref{tab:app_ft_config} and use the same setting training 100 epochs for both ViT-S and ResNet-50. Note that we use the modified SimMIM for ResNet-50 (replacing masked patches in the input image with the RGB mean) in all experiments.

As shown in Fig.~\ref{fig:in100_mask_interact} and \ref{fig:in100_mask_cl_vs_mim}, we compared MIM pre-trained models supervised methods with various augmentations and contrastive learning pre-trained methods in terms of the top-1 accuracy under various occlusion ratios. We find that MIM methods show better occlusion robustness on both Transformers and CNNs.
In addition to Sec.~\ref{sec:occlusion_robustness}, we also provide results of salient occlusion (\textit{i.e.,} dropping patches according to salient maps) for ViT-S and ResNet-50 on ImageNet-100 in Fig.~\ref{fig:app_in100_mask_salient}. Note that the occlusion ratio means the ratio of dropped and total patches, and we plot the mean of accuracy across 3 runs. Overall, we can conclude that MIM pre-trained models have stronger robustness against random and salient occlusions than supervised and contrastive-based pre-trained methods.

\subsection{Multi-order Interaction}
\label{app:interaction}
\vspace{-0.15em}
In Sec.~\ref{sec:interaction}, we interpret what is learned by MIM by multi-order interaction~\citep{deng2021discovering, zhang2020interpreting}. The interaction complexity can be represented by $I^{(m)}(i,j)$ (defined in Eqn.~\ref{eq1}), which measures the average interaction utility between variables $i,j$ on all contexts consisting of $m$ variables. Notice that the order $m$ reflects the contextual complexity of the interaction $I^{(m)}(i,j)$. For example, a low-order interaction (\textit{e.g.,} $m=0.05n$) means the relatively simple collaboration between variables $i,j$, while a high-order interaction (\textit{e.g.,} $m=0.05n$) corresponds to the complex collaboration. As figured out in the representation bottleneck \citep{deng2021discovering}, deep neural networks (DNNs) are more likely to encode both low-order interactions and high-order interactions, but often fail to learn middle-order interactions. We hypothesize that MIM helps models learn more middle-order interactions since MIM has a natural advantage in cases where some parts of the image are masked out. 
In Fig.~\ref{fig:in100_mask_interact}, we calculate the interaction strength $J^{(m)}$ (defined in Eqn.~\ref{eq2}) for fine-tuned models on ImageNet-100 using the official implementation{\footnote{\url{https://github.com/Nebularaid2000/bottleneck}}} provided by~\citet{deng2021discovering}. Specially, we use the image of $224\times 224$ resolution as the input and calculate $J^{(m)}$ on $14\times 14$ grids, \textit{i.e.,} $n=14\times 14$. And we set the model output as $f(x_S) = \log \frac{P(\hat y = y|x_S)}{1-P(\hat y = y|x_S)}$ given the masked sample $x_S$, where $y$ denotes the ground-truth label and $P(\hat y = y|x_S)$ denotes the probability of classifying the masked sample $x_S$ to the true category.

\begin{figure}[htb]
    \vspace{-1.0em}
    \centering
    \includegraphics[width=0.88\linewidth]{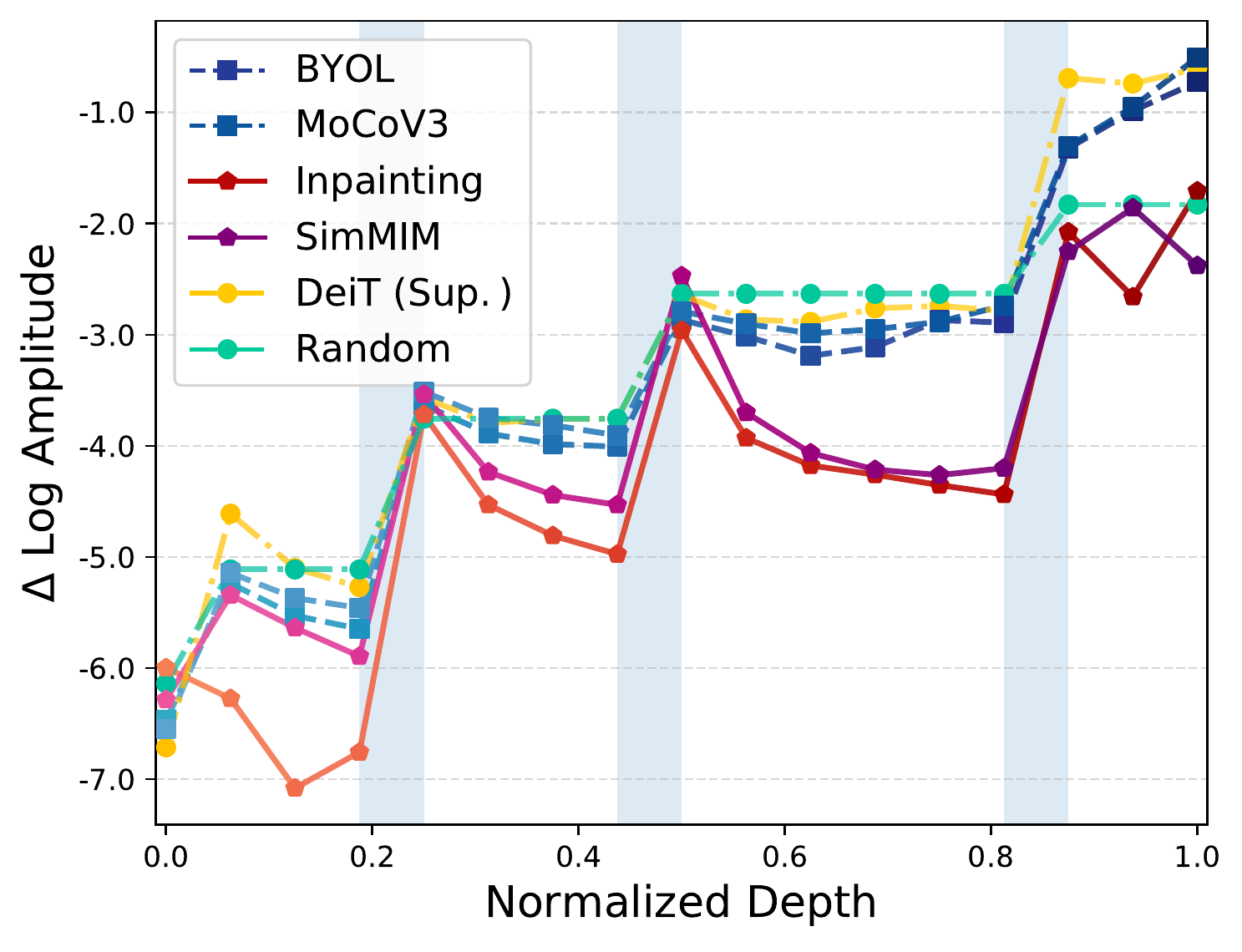}
    \vspace{-1.75em}
    \caption{Fourier transformed feature maps. The vertical axis is the relative log amplitudes of the high-frequency components, and the horizontal axis is the normalized depth of the network. The blue columns indicate the pooling layers, while the white columns indicate the convolution layers.}
    \label{fig:cnn_freq}
    \vspace{-1.0em}
\end{figure}
\begin{figure}[htb]
    \centering
    \includegraphics[width=0.88\linewidth]{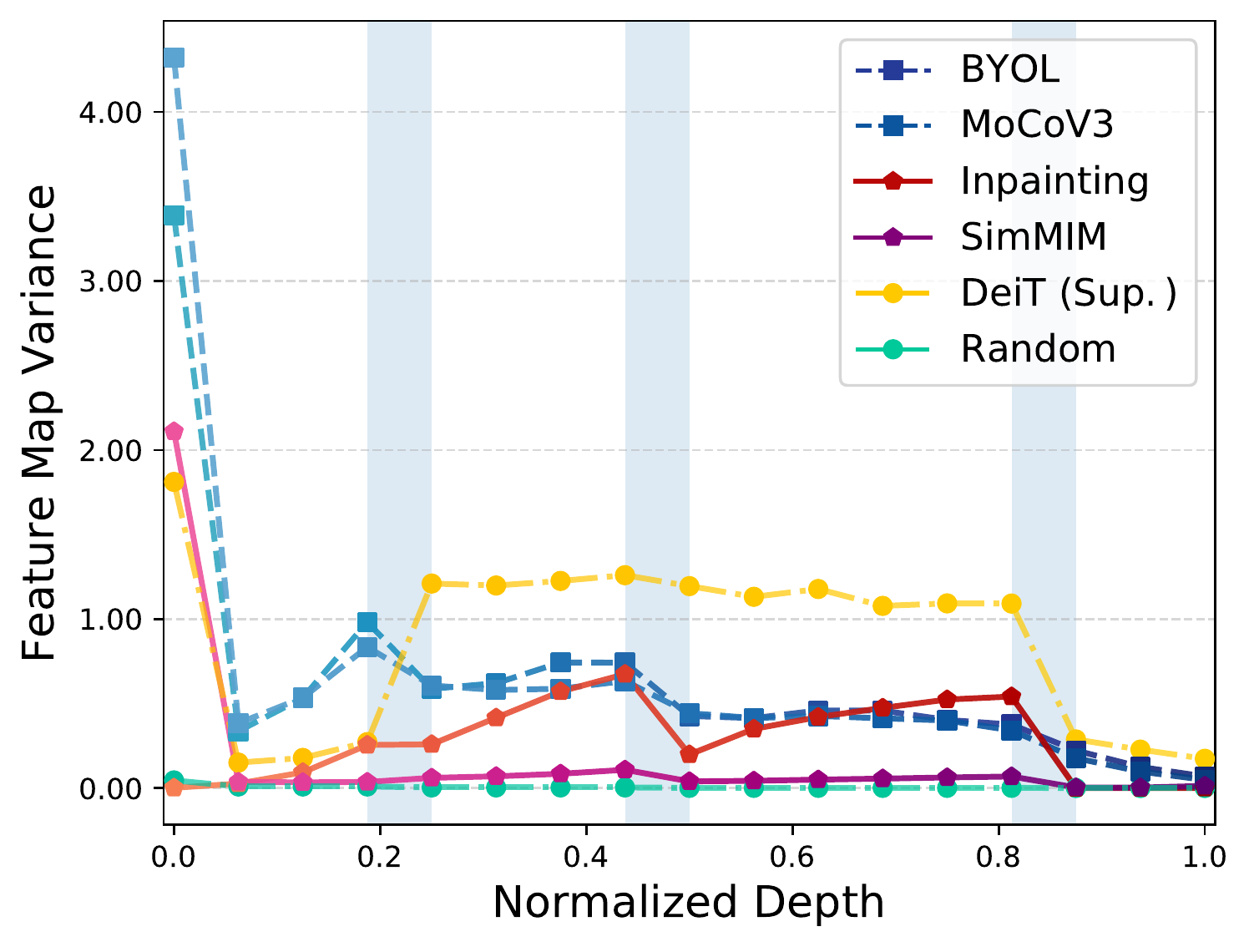}
    \vspace{-1.75em}
    \caption{Feature maps variance. The vertical axis is the average variance value of feature maps. DeiT (Sup.) is supervised pre-training. The results of the randomly initialized network are plotted for reference.}
    \label{fig:cnn_var}
    \vspace{-1.0em}
\end{figure}

\subsection{MIM from Frequency Perspective}
\label{app:analysis_freq}
We first plot the log magnitude of Fourier-transformed feature maps of ResNet-50 with different pre-training methods using the tools{\footnote{\url{https://github.com/xxxnell/how-do-vits-work}}} provided by~\citet{park2022vision} on ImageNet-1K. Following \citep{park2022vision}, we first convert feature maps into the frequency domain and represent them on the normalized frequency domain (the highest frequency components are at $\{-\pi, +\pi\}$). In Fig.~\ref{fig:cnn_freq}, we report the amplitude ratio of high-frequency components by using $\Delta \log$ amplitude. As shown in Fig. \ref{fig:cnn_freq}, inpainting and MIM show similar low-pass filtering effects at convolution layers as compared to contrastive learning. This indicates that inpainting and MIM reduce noise and uncertainty induced by high-frequency features. We argue that the reconstruction performance of MIM is mainly related to low or high-order interactions of patches~\citep{deng2021discovering}, while reconstruction performance is not directly related to the learned representation quality.
Then, we provide the standard deviation of feature maps by block depth as \citep{park2022vision,park2022blur}, which first calculates the feature map variance on the last two dimensions and then averages over the channel dimension for the whole dataset. Fig. \ref{fig:cnn_var} shows the feature variance of each layer of ResNet-50 with different pre-training methods on IN-1K. This figure indicates that MIM tends to reduce the feature map variance, and conversely, supervised training, inpainting, and contrastive learning based on CNN tend to increase variance (\textit{i.e.,} high frequencies). Compared to MIM, which learns better middle-order interactions, the inpainting task fails to filter out low-order interactions and thus leads to higher variance. To conclude, MIM methods learn middle-order interactions and reduce the feature map uncertainty (high frequencies) based on the CNN encoder for a generalized and stabilized feature extraction.

\section{More Experiment Results}
\label{app:ablation}
\subsection{Ablation of Layers for Mask Token}
\label{app:ablation_layer}
\begin{figure}[htb]
\vspace{-1.5em}
    \centering
    \caption{Ablation of the mask token in various stages (S) in ResNet-50 or layers (L) in ViT-S based on SimMIM (without $\mathcal{L}_{freq}$) on ImageNet-100.}
    \includegraphics[width=1.0\linewidth]{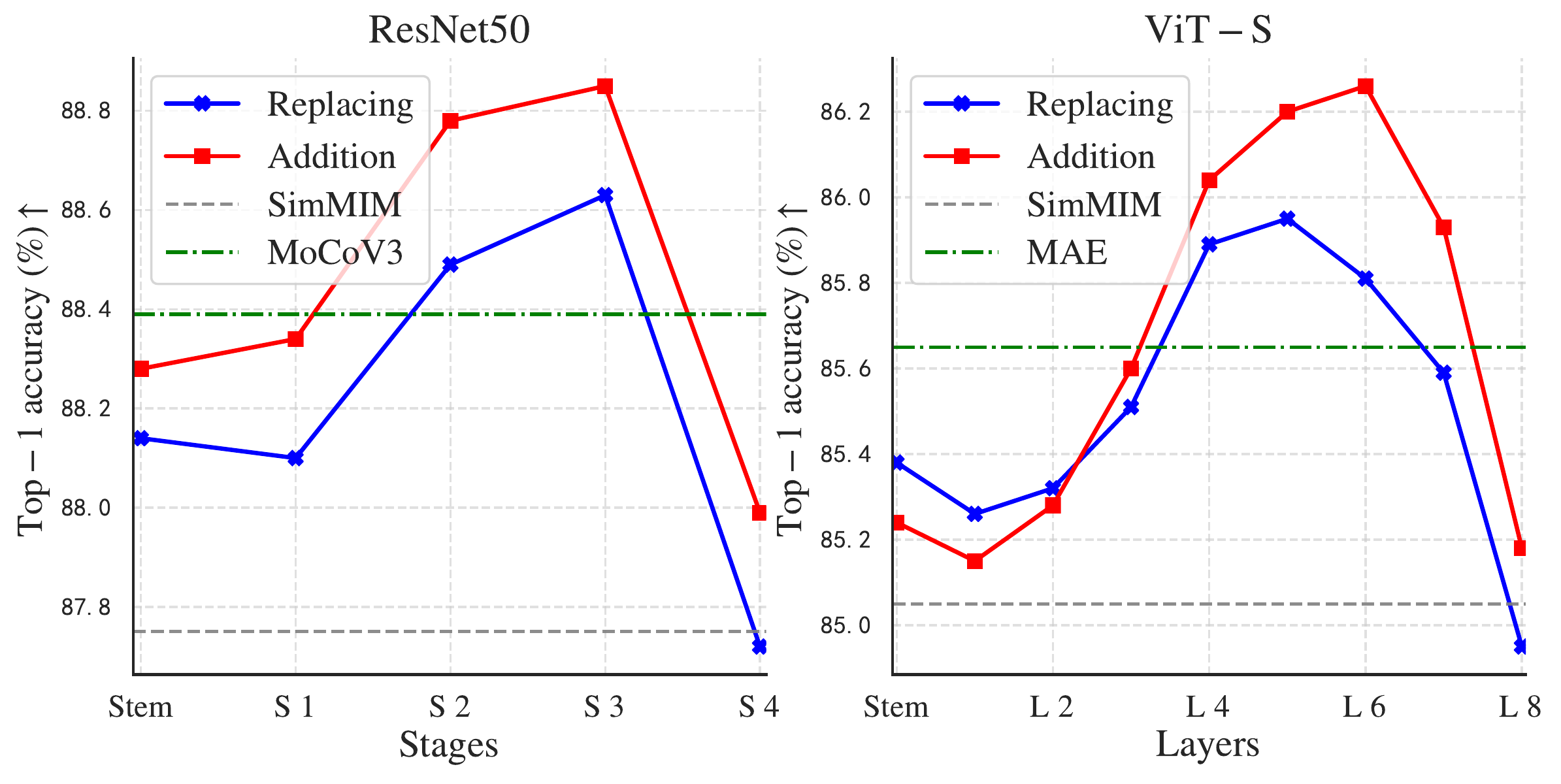}
    \label{fig:ablation_mask}
\vspace{-2.5em}
\end{figure}

In addition to Sec.~\ref{sec:exp_ablation}, we analyze the optimal stage or layer for the mask token. The ablation experiments are conducted with ResNet-50 and ViTs on IN-100 and IN-1K using the fine-tuning protocol as Sec.~\ref{sec:exp_ablation}. As shown in Fig.~\ref{fig:ablation_mask}, adding the mask token to the medium stages (stage-3 of ResNet-50) or layers (layer-5 of ViT-S) yields the best performance on the pre-trained representation. Therefore, we apply the mask token to the 3-stage or the medium layer (around 3/4 of the total layers) in A$^2$MIM by default.

\subsection{Ablation of the Proposed Modules}
\label{app:ablation_in1k}
\begin{figure*}[ht]
    \vspace{-0.5em}
\centering
    \includegraphics[width=0.98\linewidth]{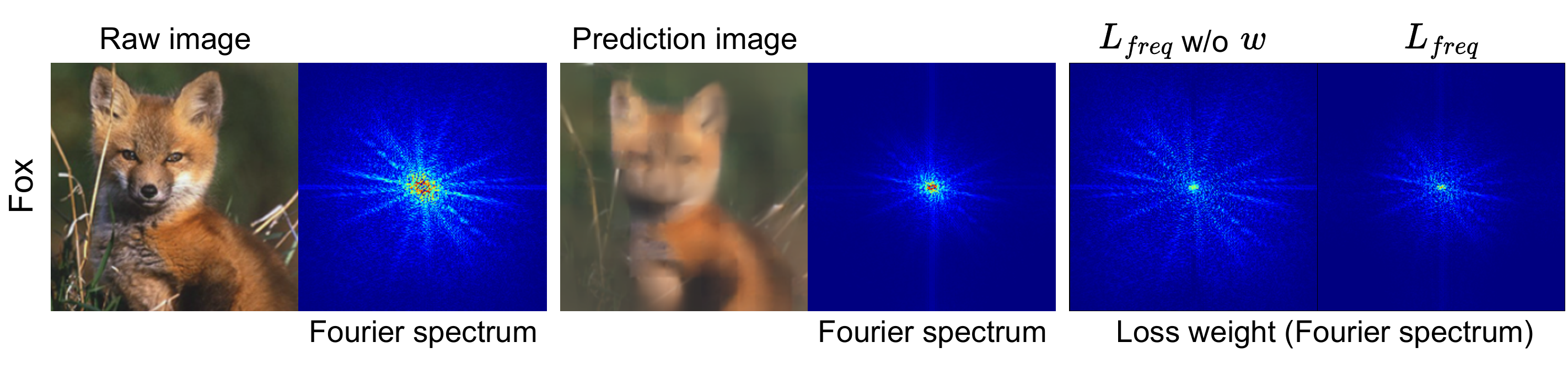}
    \vspace{-1.50em}
    \caption{
    Visualization of predicted images and $\mathcal{L}_{freq}$ loss weight in Fourier domain. From the view of the Fourier spectrum, the raw image (left) contains 99\% low-frequency components (usually present contents) and rich medium-frequency (structural patterns) and high-frequency components (local details and noises), while the predicted result (middle) provides fewer medium or high-frequency components. Calculated in the Fourier domain, the loss weights (right) of $\mathcal{L}_{freq}$ w/o $w$ help the model to learn the full spectrum while $\mathcal{L}_{freq}$ focusing on the low and medium-frequency parts, which are more likely to be low-order or middle-order interactions.
    }
    \vspace{-0.5em}
    \label{fig:app_fft_loss}
\end{figure*}

In addition to ablation studies in Sec.~\ref{sec:exp_ablation}, we provide more ablation studies and empirical analysis on the proposed $\mathcal{L}_{freq}$ in the Fourier domain, as shown in Figure~\ref{fig:app_fft_loss}. As we discussed in Sec.~\ref{sec:method}, we hypothesize that learning medium frequencies would help better learn middle-order interactions. we thereby propose $\mathcal{L}_{freq}$ to tackle the dilemma of $\mathcal{L}_{spa}$, which tends to learn low-frequency components (\textit{i.e.,} contents reflected by high-order interactions). Although the reconstruction loss in the Fourier domain has a global perception, the high-frequency components are usually constructed by local details and noises (\textit{i.e.,} low-order interactions), which might hurt the generalization abilities. Therefore, we introduce the reweight $w(u, v)$ to force the model to learn more medium-frequency components, which are identical to middle-order interactions.
Then, we perform further analysis of the masked patch size for A$^2$MIM in Tab.~\ref{tab:ablation_patch_in1k}. Note that we pre-train ResNet-50 for 100 epochs and ViT-B for 400 epochs on ImageNet-1K and report the fine-tuning results. As shown in Tab.~\ref{tab:ablation_patch_in1k}, when the mask ratio is 60\%, the optimal masked patch size is $32\times 32$ for A$^2$MIM, which is the same as SimMIM.

\begin{table}[htb]
    \vspace{-0.5em}
    \setlength{\tabcolsep}{1.0mm}
    \centering
    \caption{Ablation of masked patch size for A$^2$MIM based on ResNet-50 and ViT-B on ImageNet-1K.}
    \label{tab:ablation_patch_in1k}
\resizebox{1.0\linewidth}{!}{
    \begin{tabular}{c|cccc}
    \toprule
    Model     & Masked           & Mask  & PT    & Top-1 Accuracy (\%)                \\
              & patch size       & ratio & epoch &                                    \\ \hline
    ResNet-50 & 8~/~16~/~32~/~64 & 0.6   & 100   & 78.2~/~78.6~/~\textbf{78.8}~/~78.7 \\
    ViT-B     & 8~/~16~/~32~/~64 & 0.6   & 400   & 82.9~/~83.4~/~\textbf{83.5}~/~83.3 \\
    \bottomrule
    \end{tabular}
    }
    \vspace{-0.5em}
\end{table}

\section{Visualization Experimental Details}
\label{app:visualization}
\begin{figure*}[ht]
    \centering
    \includegraphics[width=1.0\linewidth]{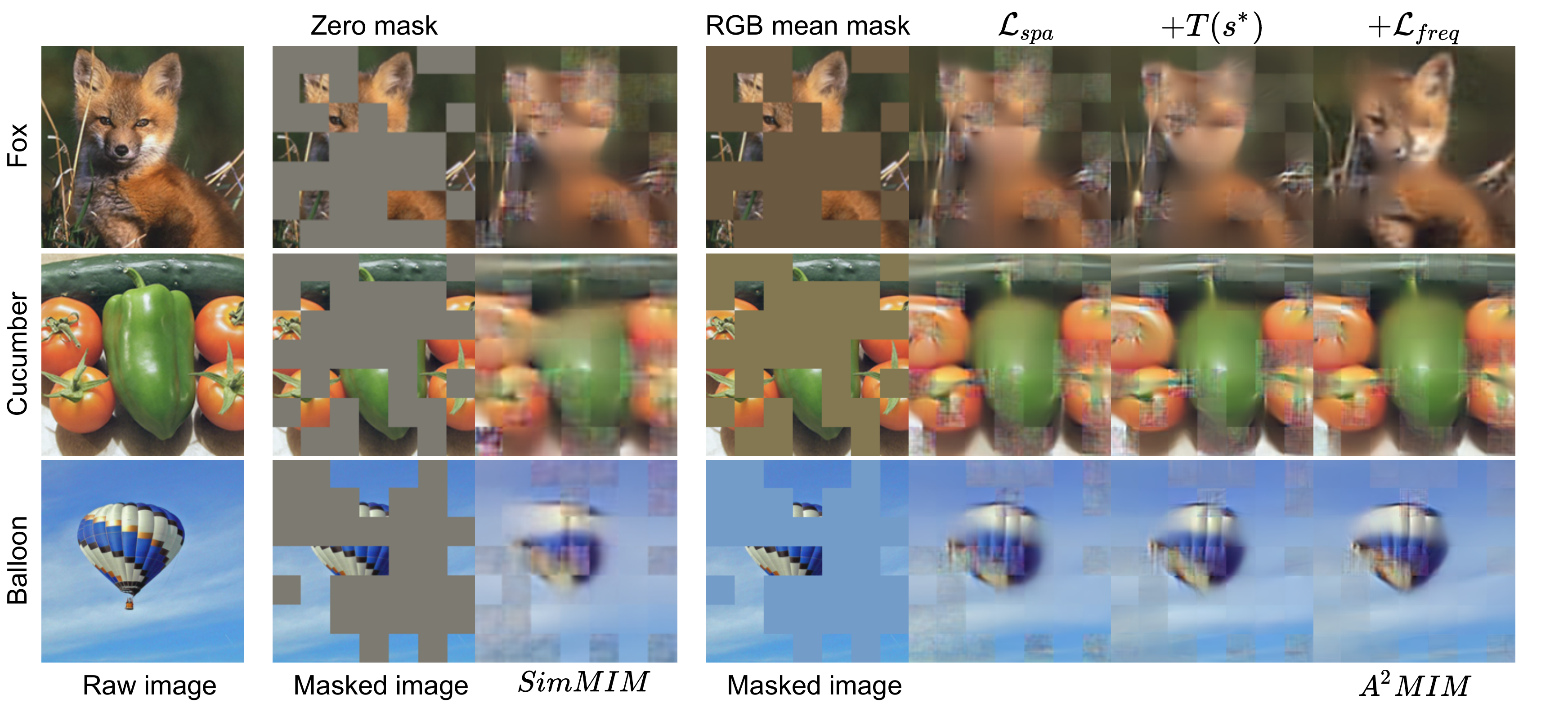}
    \vspace{-2.0em}
    \caption{Visualizations of predicted results from SimMIM (middle) and our A$^2$MIM (right) based on ResNet-50 pre-trained 100-epochs on ImageNet-1K. $T (s^*)$ denotes the mask token $T$ to the optimal stage-s in ResNet-50. We ablate the proposed components by adding them to the baseline SimMIM: replacing the zero mask with the RGB mean mask (the modified SimMIM baseline) and adding the mask token $T (s^*)$ relieve grid-like artifacts in predicted results; adding the proposed $\mathcal{L}_{freq}$ helps the model to capture more informative details.}
    \label{fig:ablation_rec_cnn}
    \vspace{-1.0em}
\end{figure*}

\begin{figure*}[ht]
    \centering
    \includegraphics[width=1.0\linewidth]{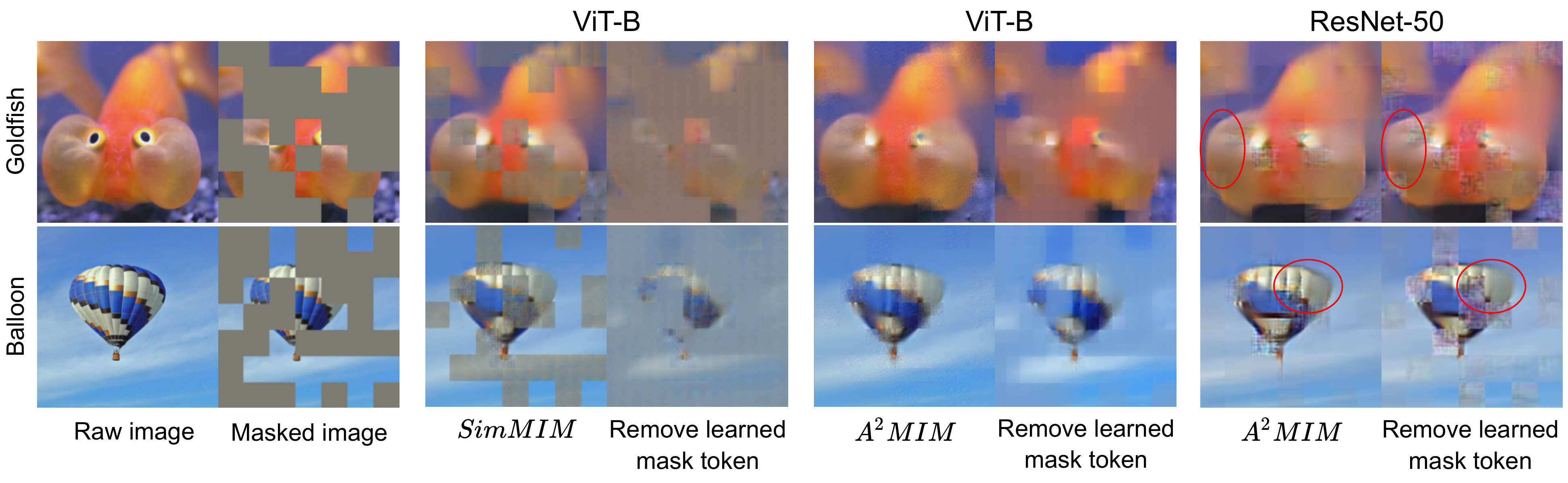}
    \vspace{-2.0em}
    \caption{Visualizations of predicted results with and without the mask token on ImageNet-1K. Notice that mask tokens are adopted in the pre-trained models based on ViT-S (300-epoch) or ResNet-50 (100-epoch). Based on ViT-S, removing the mask token corrupts both contents of masked patches and overall colors in SimMIM while only corrupting the masked contents in A$^2$MIM. Based on ResNet-50, removing the mask token slightly affects spatial details in the masked patches and causes grid-like artifacts in the unmasked patches. The different effects of the mask token in ViT-S and ResNet-50 might be because the two architectures use different spatial-mixing operators and normalization layers. As for ViTs, the self-attention operation captures informative details from unmasked patches, but the non-overlap patch embedding and layer normalization mask each patch isolated. The mask token learns the mean templates (contents) of masked patches and gathers spatial details from unmasked patches by the self-attention operation. As for CNNs, each patch shares the same contents extracted by batch normalization layers, and the convolution operation extracts features from unmasked and masked patches equally. The mask token learns more high-frequency and informative details.}
    \label{fig:rec_rm_mask_token}
    \vspace{-1.0em}
\end{figure*}

In addition to visualization results in Sec.~\ref{sec:exp_ablation}, we visualize more reconstruction results of A$^2$MIM here. Similar to Fig.~\ref{fig:ablation_rec_vit}, we ablate the proposed components in A$^2$MIM based on ResNet-50 in Fig.~\ref{fig:ablation_rec_cnn}, which demonstrates that A$^2$MIM helps ResNet-50 learn more spatial details, \textit{i.e.,} more middle-order interactions. Moreover, we study the effects of the mask token in both ViTs and CNNs in Fig.~\ref{fig:rec_rm_mask_token}.

\section{Extended Related Work}
\label{app:related_work}
In the recent decade, Deep Neural Networks (DNNs) have gained great success in various tasks with full supervision, such as computer vision~\cite{he2016deep,iccv2021Swin, 2017iccvmaskrcnn, icml2023song}, natural language processing~\cite{vaswani2017attention,devlin2018bert, Radford2018GPT1}, and graph representation learning~\cite{iclr2019gin,icml2023ExplainingGNN}. As DNNs scale up with more parameters, pre-training without labels by leveraging pre-text tasks has become increasingly popular. In addition to Sec.~\ref{sec:related}, we provide extended discussions of two types of popular self-supervised vision pre-training approaches.

\paragraph{Contrastive Learning.}
Contrastive learning learns instance-level discriminative representations by extracting invariant features over distorted views of the same data, which is first introduced by CPC~\citep{Oord2018cpc}, CMC~\citep{eccv2020CMC}, and NPID~\citep{cvpr2018npid}. MoCo~\citep{cvpr2020moco} and SimCLR~\citep{chen2020simple} adopted different mechanisms to introduce negative samples for contrast with the positive. BYOL~\citep{nips2020byol} and its variants~\citep{chen2020simsiam,nips2021revitalizing} further eliminate the requirement of negative samples to avoid representation collapse. Besides pairwise contrasting, SwAV~\citep{caron2020unsupervised} clusters the data while enforcing consistency between multi-augmented views of the same image. Barlow Twins~\citep{zbontar2021barlow} proposed to measure the cross-correlation matrix of distorted views of the same image to avoid representation collapsing. Meanwhile, some efforts have been made on top of contrastive methods to improve pre-training quality for specific downstream tasks~\citep{iccv2021detco, xiao2021region,cvpr2021casting,wu2021align}, which conduct fine-grained contrastive supervisions. MoCo.V3~\citep{chen2021empirical} and DINO~\citep{iccv2021dino} adopted ViT~\citep{dosovitskiy2020image} in self-supervised pre-training to replace CNN backbones.

\paragraph{Autoregressive Modeling.}
Autoencoders (AE) is a typical type of network architecture that allows representation learning with no annotation requirement~\citep{hinton1993autoencoders}. By forcing denoising property onto the learned representations, denoising autoencoders \citep{vincent2008extracting, vincent2010stacked} are a family of AEs that reconstruct the uncorrected input signal with a corrupted version of the signal as input.
Generalizing the notion of denoising autoregressive modeling, masked predictions attracted the attention of both the NLP and CV communities. BERT~\citep{devlin2018bert} performs masked language modeling (MLM), where the task is to classify the randomly masked input tokens. Representations learned by BERT as pre-training generalize well to various downstream tasks. For CV, inpainting tasks~\citep{pathak2016context} to predict large missing regions using CNN encoders and colorization tasks~\citep{eccv2016coloring} to reconstruct the original color of images with removed color channels are proposed to learn representation without supervision. With the introduction of Vision Transformers (ViTs) \citep{dosovitskiy2020image,iccv2021Swin}, iGPT~\citep{chen2020generative} predicts succeeding pixels given a sequence of pixels as input. MAE~\citep{he2021masked} and BEiT~\citep{bao2021beit} randomly mask out input image patches and reconstruct the missing patches with ViTs. Compared to MAE, MaskFeat~\citep{wei2021masked} and SimMIM~\citep{xie2021simmim} adopt linear layers as the decoder instead of another Transformer as in MAE. MaskFeat applied HOG as the prediction target instead of the RGB value. Other research endeavors \citep{el2021large,zhou2021ibot,assran2022masked, nips2021vatt,2022dilemma} combine the idea of contrastive learning (CL) with MIM. 
SplitMask~\citep{el2021large} proposed to use half of the image pixels to predict the other half while applying InfoNCE loss~\citep{van2018representation} across the corresponding latent features. MSN~\citep{assran2022masked} matches the representation of an image view containing randomly masked patches and the original unmasked image. Similarly, iBOT~\citep{zhou2021ibot} adopts the Siamese framework to combine self-distillation with MIM.
Moreover, Data2Vec~\citep{baevski2022data2vec} proposed a framework that applies the masked prediction idea for either speech, NLP, or CV.
However, most MIM works are confined to ViT architectures, recently proposed CIM \citep{fang2022corrupted} uses the output of a pre-trained tokenizer as the target and takes the output of a frozen BEiT as the encoder's input as a workaround to enable MIM on CNNs.

In this work, we propose A$^2$MIM with no components native to ViTs adopted to perform MIM with ViTs and CNNs. Two concurrent two after A$^2$MIM, SparK~\citep{iclr2023spark} and ConvNeXt.V2~\citep{Woo2023ConvNeXtV2}, designed CNN-based MIM with sparse convolutions to tackle the irregular masked images. Compared to them, A$^2$MIM provides empirical explanations of why MIM works and designs an architecture-agnostic framework.


\end{document}